\documentclass{article} 

\usepackage{amsmath,amssymb,amsthm,amsfonts}
\usepackage{caption,subcaption}
\usepackage{times}
\usepackage{xr}
\usepackage{hyperref}
\usepackage{url}
\usepackage{amssymb}
\usepackage[round]{natbib}
\usepackage{amsbsy}
\usepackage{amsthm}
\usepackage{amsmath}
\usepackage{paralist}
\usepackage{bm}
\usepackage{booktabs}
\usepackage{rotating}
\usepackage{cases}
\usepackage{mathtools}
\usepackage{graphicx}
\usepackage{setspace}
\usepackage{textcomp}
\usepackage[accepted]{arxivicml2016}

\newcommand{\vnorm}[1]{\left|\left|#1\right|\right|}

\icmltitlerunning{Smoothed Dyadic Partitioning}

\begin{document}

\twocolumn[
\icmltitle{Deep Nonparametric Estimation of Discrete Conditional Distributions via Smoothed Dyadic Partitioning}

\icmlauthor{Wesley Tansey}{tansey@cs.utexas.edu}
\icmladdress{Department of Computer Science,
            University of Texas at Austin}
\icmlauthor{Karl Pichotta}{kpich@cs.utexas.edu}
\icmladdress{Department of Computer Science,
            University of Texas at Austin}
\icmlauthor{James G.~Scott}{james.scott@mccombs.utexas.edu}
\icmladdress{Department of Information, Risk, and Operations Management; 
            Department of Statistics and Data Sciences,
            University of Texas at Austin}

\icmlkeywords{conditional probability, distribution estimation, deep learning, neural networks}

\vskip 0.3in
]

\begin{abstract}
We present an approach to deep estimation of discrete conditional probability distributions. Such models have several applications, including generative modeling of audio, image, and video data. Our approach combines two main techniques: dyadic partitioning and graph-based smoothing of the discrete space. By recursively decomposing each dimension into a series of binary splits and smoothing over the resulting distribution using graph-based trend filtering, we impose a strict structure to the model and achieve much higher sample efficiency. We demonstrate the advantages of our model through a series of benchmarks on both synthetic and real-world datasets, in some cases reducing the error by nearly half in comparison to other popular methods in the literature. All of our models are implemented in Tensorflow and publicly available at \href{https://github.com/tansey/sdp}{this url}.
\end{abstract}

\section{Introduction}
\label{sec:introduction}
Recently there has been a flurry of interest in using deep-learning methods to estimate conditional probability distributions. The applications of such models cover a wide variety of scientific areas, from cosmology \cite{ravanbakhsh:etal:2016} to health care \cite{ranganath:etal:2016,ng:etal:2017}. A subset of this area deals specifically with discrete conditional distributions, where an explicit estimation of the likelihood is desired---as opposed to simply the ability to sample the distribution, as with GAN-based models \cite{goodfellow:etal:2014}. Deep learning models that output discrete probability distributions have achieved state-of-the-art results in text-to-speech synthesis \cite{oord:etal:2016:wavenet}, image generation \cite{oord:etal:2016:pixelcnn,oord:etal:2016:pixelrnn,oord:etal:2016:conditionalpixelcnn,gulrajani:etal:2016:pixelvae,salimans:etal:2017:pixelcnnpp}, image super resolution \cite{dahl:etal:2017}, image colorization \cite{deshpande:etal:2016}, and EHR survival modeling \cite{ranganath:etal:2016}. Methodological improvements to deep discrete conditional probability estimation (CPE) therefore have the potential to make substantial improvements in many fields of interest in machine learning.

In this paper we focus on the specific form of the deep CPE model used when estimating low-dimensional data such as audio waveforms (1d) or pixels (3d). Typically, this is the output layer of a deep neural network and represents either the logits of a multinomial distribution or the parameters of a mixture model, such as a Gaussian mixture model (also known as a mixture density network \cite{bishop:mdn:1994}). Previous work \cite{oord:etal:2016:wavenet,oord:etal:2016:pixelcnn} has found empirically that using a multinomial model often outperforms GMMs on discrete data. Methods to improve performance over the naive multinomial model often involve sophisticated compression of the space into a smaller number of bins \cite{oord:etal:2016:wavenet} or hand-crafting a mixture model to better-suit the marginal distribution of the data \cite{salimans:etal:2017:pixelcnnpp}.

We propose an alternative model, Smoothed Dyadic Partitions (SDP), as a drop-in replacement for these conventional, widely used CPE approaches. SDP performs a dyadic decomposition of the discrete space, transforming the probability mass estimation into a series of left/right split estimations. During training, SDP locally smooths the area around the target value in the discrete space using a graph-based smoothing technique. These two techniques seek to leverage the inherent spatial structure in the discrete distribution to enable points to borrow statistical strength from their nearby neighbors and consequently improve the estimation of the latent conditional distribution.

SDP out-performs both multinomial and mixture models on synthetic and real-world datasets.  These empirical results show that SDP is unique in its ability to provide a flexible fit that smooths well without suffering from strong inductive bias, particularly at the boundaries of the  space (e.g.~pixel intensities 0 and 255 in an image problem).  Our SDP design also involves specific attention to efficient implementation on GPUs.  These design choices, together with a local neighborhood sampling scheme for evaluating the regularizer, ensure that the approach can scale to large (finely discretized) domains in low-dimensional space.

The remainder of this paper is organized as follows. Section \ref{sec:dyadic_partitioning} outlines our dyadic partitioning strategy. Section \ref{sec:smoothing} details our approach to smoothing the underlying discrete probability space using graph-based trend filtering. Section \ref{sec:experiments} presents our experimental analysis of SDP and our benchmarks confirming its strong performance. Section \ref{sec:discussion} provides a discussion of related work and the limitations of our model. Finally, Section \ref{sec:conclusion} gives concluding remarks.


\section{Dyadic partitioning of the discrete space}

\label{sec:dyadic_partitioning}

Our model relies on representing the target discrete distribution using a tree rather than a flat grid. Tree-based models for distribution estimation have a long history of success in machine learning. This includes seminal work using \textit{k-d} trees for nonparametric density estimation \cite{gray:moore:2003} and hierarchical softmax for neural language models \cite{morin:bengio:2005}. More recent work includes, for instance, spatial discrete distribution estimation of background radiation \cite{tansey:scott:2016:multiscale}. We draw inspiration from these past works in the design of our SDP model.

\subsection{Dyadic decomposition}
\label{subsec:dyadic:decomposition_1d}

Rather than outputting the logits of a multinomial distribution directly, we instead create a balanced binary tree with its root node in the center of the discrete space. From the root, we recursively partition the space into a series of half spaces, resulting in $n-1$ nodes for a discrete space of size $n$. The deep learning model then outputs the splitting probabilities for every node, $\mathcal{E}_i$, parameterized as the logits in a series of independent binary classification tasks,
\begin{equation}
p(y > b_i | x) = \frac{1}{1 + \exp(-\mathcal{E}_i)} \, ,
\label{eq:node_probs}
\end{equation}
where $b_i$ is the center of the node.

\begin{figure}
\centering
\includegraphics[width=0.45\textwidth]{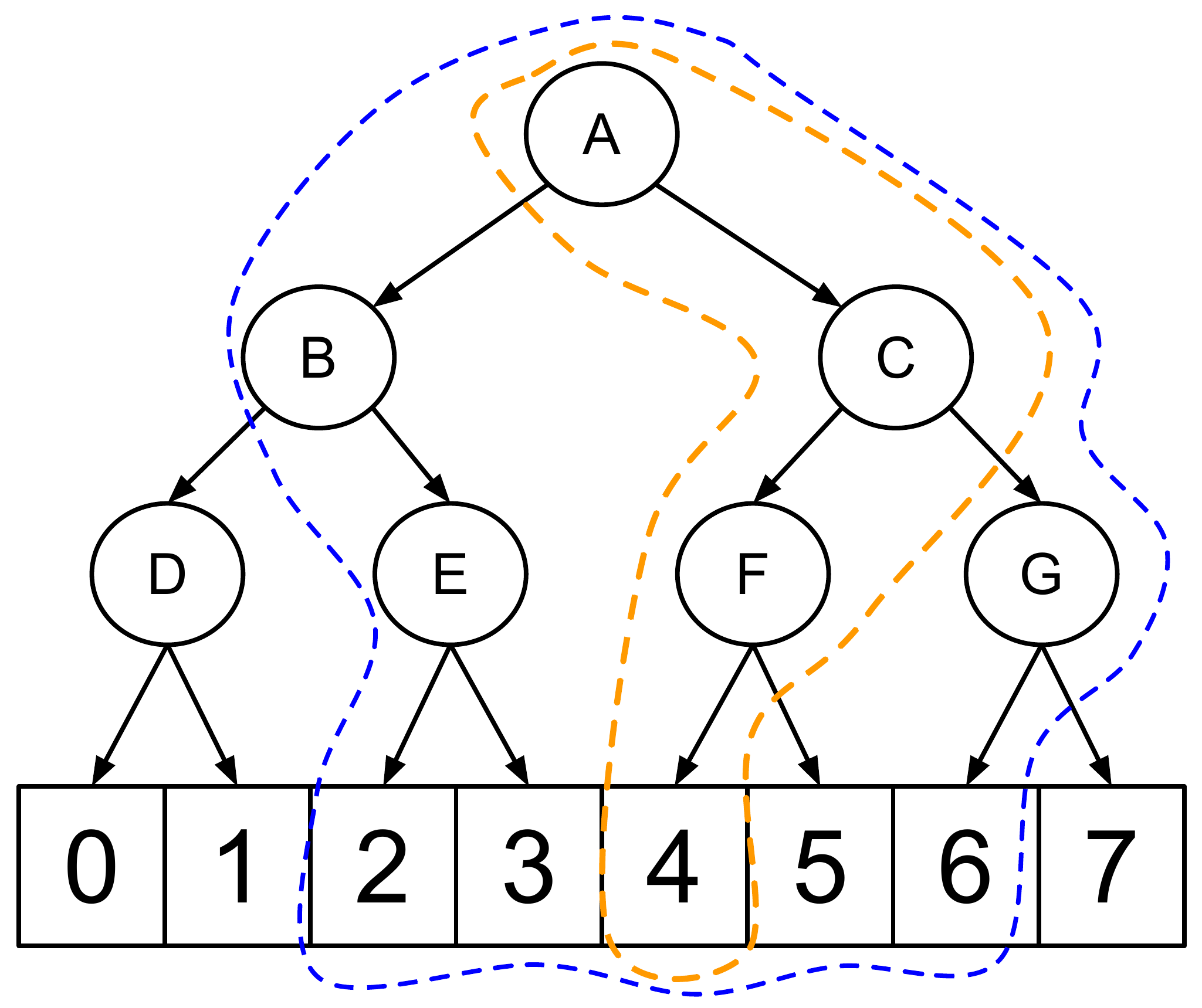}
\caption{\label{fig:sdp_algorithm} An illustration of our algorithm. The discrete space is recursively partitioned into a series of binary left-right splits and the model outputs the splitting probability for each node. During training, computing the log probability of a target label only requires calculating the nodes path to the label in the tree and its local smoothing neighborhood. In the example above, the target label is 4 and the neighborhood radius is 2, resulting in the need to calculate the target path (orange) and the paths of the surrounding 2 labels on each side (blue). As the size of the discrete space grows larger, and especially in multi-dimensional spaces, the computational savings of this approach become substantial.}
\end{figure}

Figure \ref{fig:sdp_algorithm} presents an illustration of the dyadic partitioning (DP) approach. For ease of exposition, we denote the conditional probabilty of y being greater than the node value as simply $p(N)$ for a given node $N$. For a target value of $y_i = 4$ with some training example $x_i$, we calculate the log probability during training as $\text{log}(p(y_i = 4 | x_i)) = \text{log}\left[p(A)(1-p(C))(1-p(F))\right]$. The training objective for the model is then the sum of the log probabilities of the training data.

There are several computational advantages to using this particular structure compared to a multinomial. For large spaces, multinomial models typically require some form of negative sampling \cite{mikolov:etal:2013:word2vec,jean:etal:2014} at training time to remain efficient. In the DP model, however, every split is conditionally independent of the rest of the tree and there is no partition function to estimate. Instead, we only require the $\mathcal{O}(\text{log}_2n)$ path from the root to the target node to estimate the probability of a given training example. Using a balanced tree also guarantees that every path has a fixed-length of $\lceil\text{log}_2n\rceil$, making vectorization on a GPU straightforward. Finally, because each node is only dealing with splitting its local region, the resulting computations are much more numerically stable and rarely result in very small or large log-probabilities-- a problem that often plagues both multinomial and mixture model approaches.

\subsection{Multiple dimensions}
\label{subsec:dyadic:multidim}

We extend the DP approach to multi-dimensional distributions in a manner similar to a balanced \textit{k-d} tree. We enumerate the splits in the tree in a breadth-first fashion, alternating dimensions at each level of the tree. This has two distinct advantages over a depth-first approach of enumerating the first dimension before proceeding to the next dimension. The breadth-first approach means that all nodes close in euclidean space will share more coarse-grained parents. This makes training more efficient by imposing a more principled notion of structure on the discrete space. It also improves computational efficiency for the local smoothing strategy described in Section \ref{subsec:smoothing:local}, as nearby values have heavily-overlapping paths; this results in a well-utilized GPU cache when training.

\section{Smoothing via graph-based trend filtering}

\label{sec:smoothing}

The DP approach described above imposes a spatial structure on the discrete space. In the example from Figure \ref{fig:sdp_algorithm}, an example with $y_i = 4$ is likely to result in an increase in probability of $p(y_i = 5 | x_i)$ as well, since both $A$ and $C$ will increase in the direction of $5$ and only $F$ will be downweighted. However, it will clearly \textit{decrease} the likelihood of $p(y_i = 3 | x_i)$, since it will shift the probability of $A$ away from $3$ and leave the other nodes in the path of $3$ unchanged. This imabalance in updates is likely to lead to jagged estimations of the underlying conditional distribution. To address this issue, we incorporate a smoothing regularizer into SDP that spreads out the probability mass to nearby neighbors as a function of distance in the underlying discrete space, rather than only their specific DP paths.

\subsection{Trend filtering logits}
\label{subsec:smoothing:trendfiltering}
Trend filtering \cite{kim:etal:2009,tibshirani2014adaptive} is a method for performing adaptive smoothing over discrete spatial lattices. In the univariate case with a Gaussian loss, the solution to the trend filtering minimization problem results in a piecewise polynomial fit similar to a spline with adaptive knot placement. The order of the polynomial is a hyperparameter chosen to minimize some objective criterion such as AIC or BIC; in SDP, we use validation error. Recent methods \cite{wang:etal:2016} extend trend filtering to arbitrary graphs and theoretical results show that trend filtering has strong minimax rates \cite{sadhanala:etal:2016} and is optimally spatially adaptive for univariate discrete spaces \cite{guntuboyina:etal:2017}.

To smooth the conditional distributions, we employ a graph-based trend filtering penalty applied to the conditional log-probabilities output by our model. This yields a regularized loss function for the $i^{\text{th}}$ training sample,
\begin{equation}
\mathcal{L}_i = -\text{log}\left[p(y = y_i | x_i)\right] + \lambda \vnorm{\Delta^{(k)}\texttt{vec}(\text{log}\left[p(y | x_i)\right])}_1 \, ,
\label{eq:trend_filtering_full}
\end{equation}
where $\Delta^{(k)}$ is the sparse $k^{\text{th}}$-order graph trend filtering penalty matrix as defined in \cite{wang:etal:2016} and \texttt{vec} is a function mapping the $d$-dimensional discrete conditional distribution over $y$ to a vector. The $\lambda$ term is a hyperparameter that controls the tradeoff between the fit to the training data and the smoothness of the underlying distribution. As we show in Section \ref{subsec:experiments:synthetic}, as the size of the dataset increases, the best SDP setting will drift towards smaller values of $\lambda$. Thus, in small-sample regimes SDP relies on trend filtering to smooth out the underlying space, whereas in large-sample regimes it converges to the pure DP model.

\subsection{Local smoothing via neighborhood sampling}
\label{subsec:smoothing:local}
A naive implementation of the trend filtering regularizer would require evaluating all the nodes in the discrete space. This would remove many of the computational performance advantages of the DP model described in Section \ref{subsec:dyadic:decomposition_1d}. To ensure that SDP scales to large spaces, we smooth only over a local neighborhood around the target value. Specifically, for a given $y_i$, we smooth over all nodes in the hypercube of radius $r$ centered at $y_i$. The resulting regularization loss is then only over this subset of the space,
\begin{equation}
\mathcal{L}_i = -\text{log}\left[p(y = y_i | x_i)\right] + \lambda \vnorm{\tilde{\Delta}^{(k)}\ell(y_i, x_i)}_1 \, .
\label{eq:trend_filtering_local}
\end{equation}
In \eqref{eq:trend_filtering_local}, $\tilde{\Delta}^{(k)}$ is the graph trend filtering matrix for a discrete grid graph of size $(2r+1)^d$ and $\ell(\cdot)$ is the neighborhood selection function that returns the vector of local conditional logits to smooth. Figure \ref{fig:sdp_algorithm} provides an illustration of the local sampling for a neighborhood radius of size 2 and a target label of 4. 

By only needing to compute the values of a local neighborhood, the SDP model regains its computational efficiency. For instance, in the case of a neighborhood radius of size 5 in a 3d scenario where each dimension is of size 64, the full smoothing model would have to calculate $\approx 262K$ output DP nodes. The local smoother on the other hand only needs paths of size 24 for 1331 logits for an upper bound of $\approx 32K$ nodes. Even though this is already a sharp reduction ($\approx 88\%$), most of the local neighborhood will have highly-overlapping paths and thus the average number of nodes sampled is much lower than the upper bound.

\section{Experiments}
\label{sec:experiments}
We evaluate SDP on a series of benchmarks against real and synthetic data. First, we show how the dyadic partitioning is effected by the trend filtering with different neighborhood sizes. We then compare SDP against approaches found in the recent literature and highlight the particular pathologies of each method. Finally, we measure the performance of each method on real datasets of one, two, and three-dimensional discrete conditional target distributions.

\subsection{Neighborhood size}
\label{subsec:experiments:neighborhoods}

As noted in Section \ref{subsec:smoothing:local}, the local smoothing strategy of SDP introduces a new hyperparameter to tune. To evaluate the effect of different choices of neighborhood size we consider a toy example of performing marginal density estimation on a large 1000-bin 1d grid. We use a piecewise-linear ground truth function to parameterize the logits of the true distribution:
\begin{equation}
\mathcal{E}_i = \left\{\begin{array}{ll}
                        0.5 & \text{if } i = 1 \\
                        0.5 + \mathcal{E}_{i-1} & \text{if } 1 < i \le 300 \\
                        -2 + \mathcal{E}_{i-1} & \text{if } 300 < i \le 450 \\
                        0.9 + \mathcal{E}_{i-1} & \text{if } 450 < i \le 750 \\
                        0.5 + \mathcal{E}_{i-1} & \text{if } 750 < i \le 850 \\
                        -1 + \mathcal{E}_{i-1} & \text{if } 850 < i \le 1000
                        \end{array}\right. \, .
\label{eq:experiments:neighborhood_truth}
\end{equation}
We then standardize the logits and draw 5000 samples from the corresponding multinomial.

\begin{figure}
\centering
\includegraphics[width=0.45\textwidth]{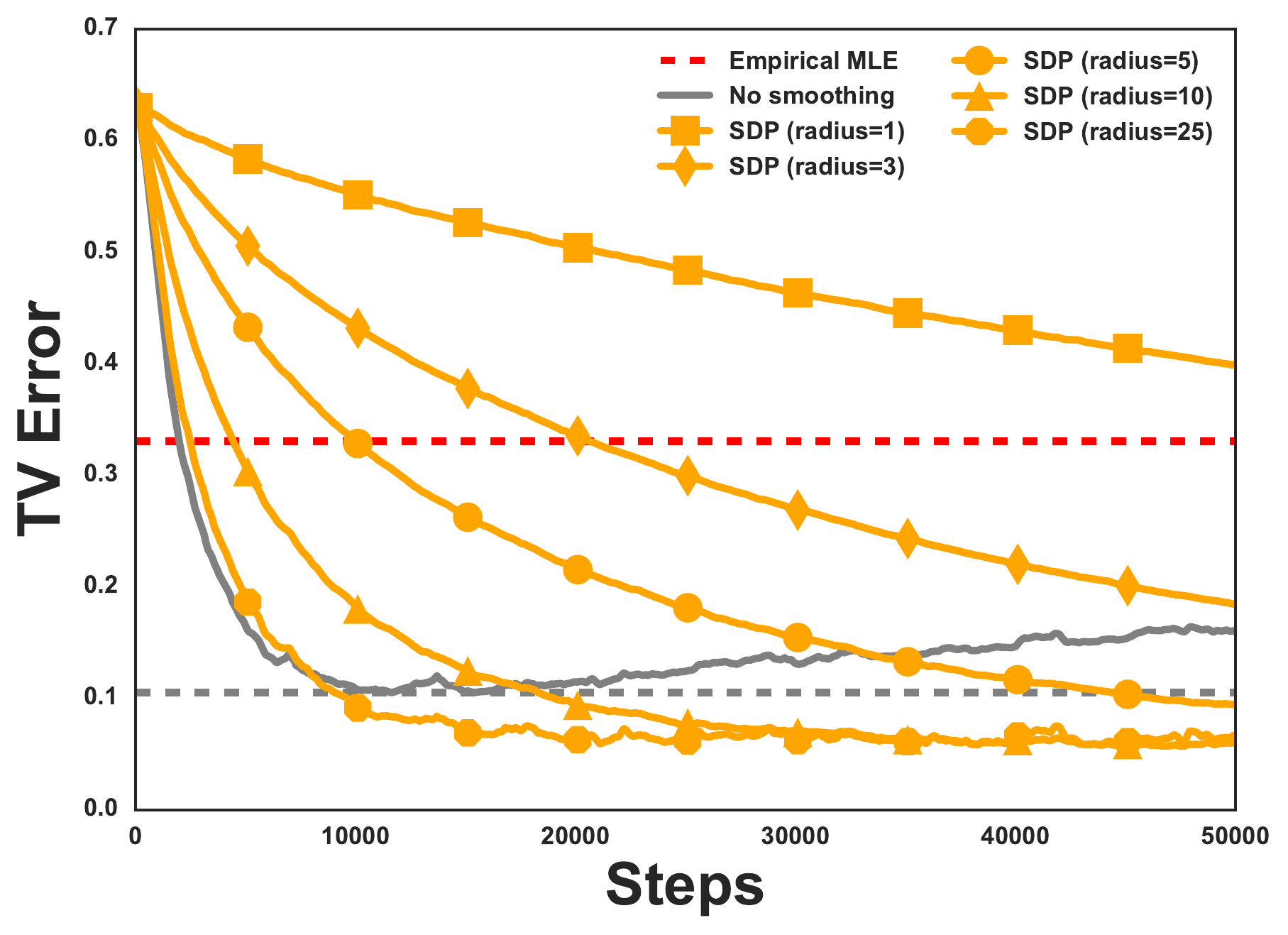}
\caption{\label{fig:neighborhoods_error} Learning rate plots for different neighborhood sizes on the illustrative example in Section \ref{subsec:experiments:neighborhoods}. As the local smoothing area increases, the model becomes more sample efficient at the cost of being more computationally expensive. The solid gray line shows the performance of the unsmoothed model; the dashed gray line is the peak performance for the unsmoothed model. After a rapid learning process, the unsmoothed model begins to overfit and starts to return to the empirical distribution. In contrast, the larger neighborhoods quickly converge to an estimate better than the unsmoothed model ever achieves.}
\end{figure}

As a baseline, we consider an unsmoothed dyadic partitioning model which simply performs unregularized maximum likelihood estimation (MLE). We then evaluate SDP with five different neighborhood radius sizes: 1, 3, 5, 10, and 25. We fix the other settings to use first-order ($k=1$) trend filtering with the $\lambda$ penalty fixed at $0.02$. All models are fit via Adam with a learning rate of $10^{-2}$, $\epsilon$ of $0.1$, and mini-batch size of $10$. We run the experiment for 50K steps and plot the total variation (TV) error between the true distribution and the estimated distribution in Figure \ref{fig:neighborhoods_error}.

The baseline unsmoothed method (gray solid line) quickly reaches an error of around $0.1$ (gray dashed line), much lower than the empirical MLE (dashed red line). However, as the model continues to train it begins to overfit and will eventually converge to the empirical MLE if training is allowed to continue. The smoothing in SDP acts as a regularizer which prevents this overfitting, as is seen in the case of the larger radii of size 10 and 25. These models converge nearly as quickly as the unsmoothed model, but both reach a better TV error and do not begin to overfit. 

When the neighborhood size is small, as in the case of the radii of size 1 and 3, the smoothing can substantially slow down learning. On the other hand, wall-clock time for SDP scales linearly with the neighborhood size. This creates a clear tradeoff for SDP: smaller neighborhoods are computationally more efficient, but may require many more samples to converge. Fortunately, a radius of 25 is still only 5\% of the total size of the grid, yielding a considerable wall-clock speedup over the full trend filtering penalty. We found that neighborhoods larger than 25 did not yield any additional sample efficiency nor asymptotic performance benefits on this experiment. It may also be possible to employ a hybrid approach of fitting an unsmoothed model initially, then smoothing later, though we have not explored such an approach.

\subsection{Synthetic conditional distributions}
\label{subsec:experiments:synthetic}

We next create a synthetic benchmark to evaluate the sample efficiency and systematic pathologies of both our method and other methods used in the recent literature. Our task is a variant on the well-known MNIST classification problem but with the twist that rather than mapping each digit to a latent class, each digit is mapped to a latent discrete distribution. For each sample image, we generate a label ($y$) by first mapping the digit to its corresponding distribution and then sampling $y$ as a draw from that distribution, resulting in a training set of  (X, y) values where X is an image (whose digit class is not explicitly known by the model) and y is an integer.

\begin{figure*}[!ht]
\centering
\begin{subfigure}[t]{0.45\textwidth}
\includegraphics[width=\textwidth]{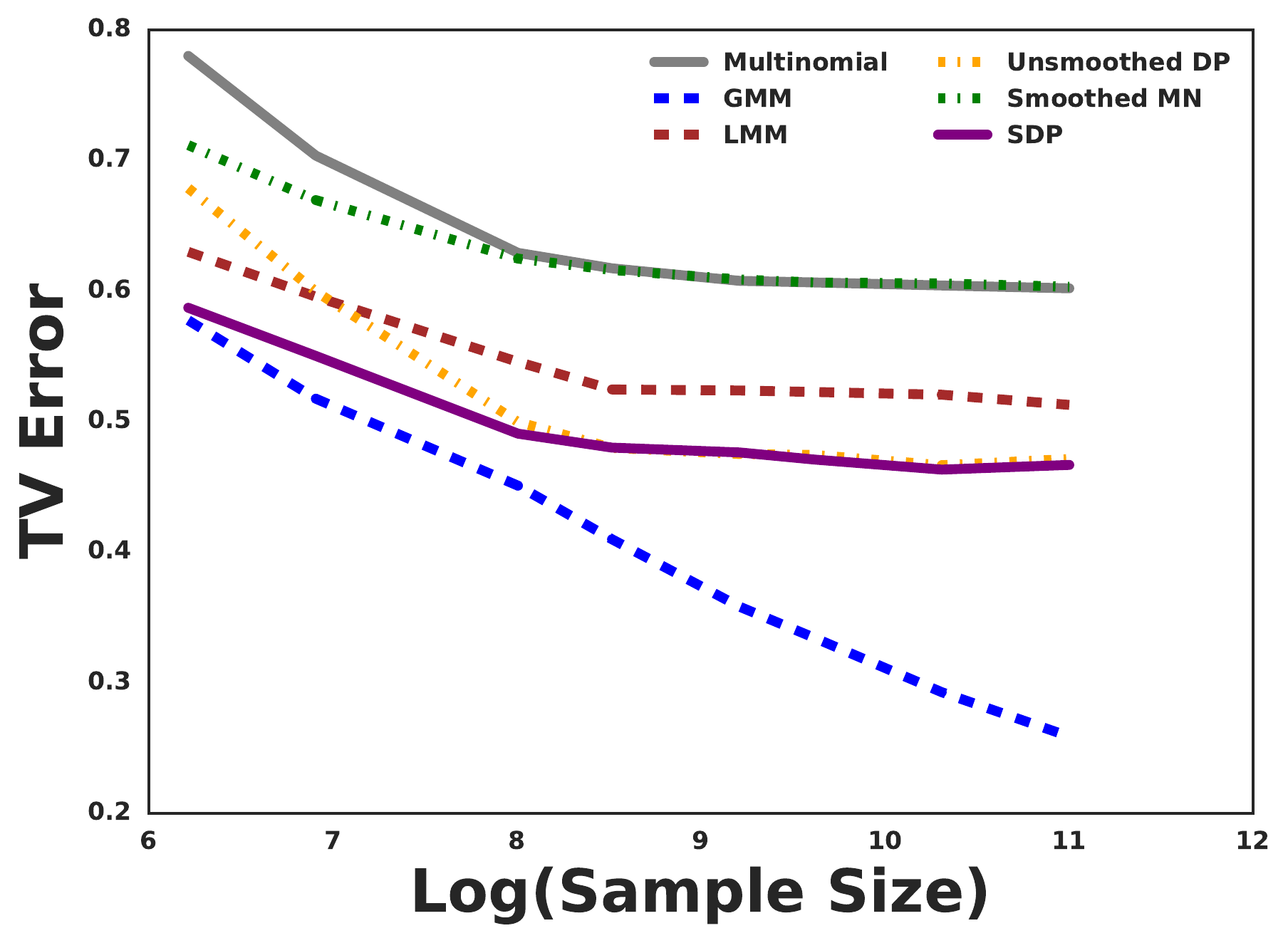}
\caption{GMM Distribution}
\end{subfigure}
\begin{subfigure}[t]{0.45\textwidth}
\includegraphics[width=\textwidth]{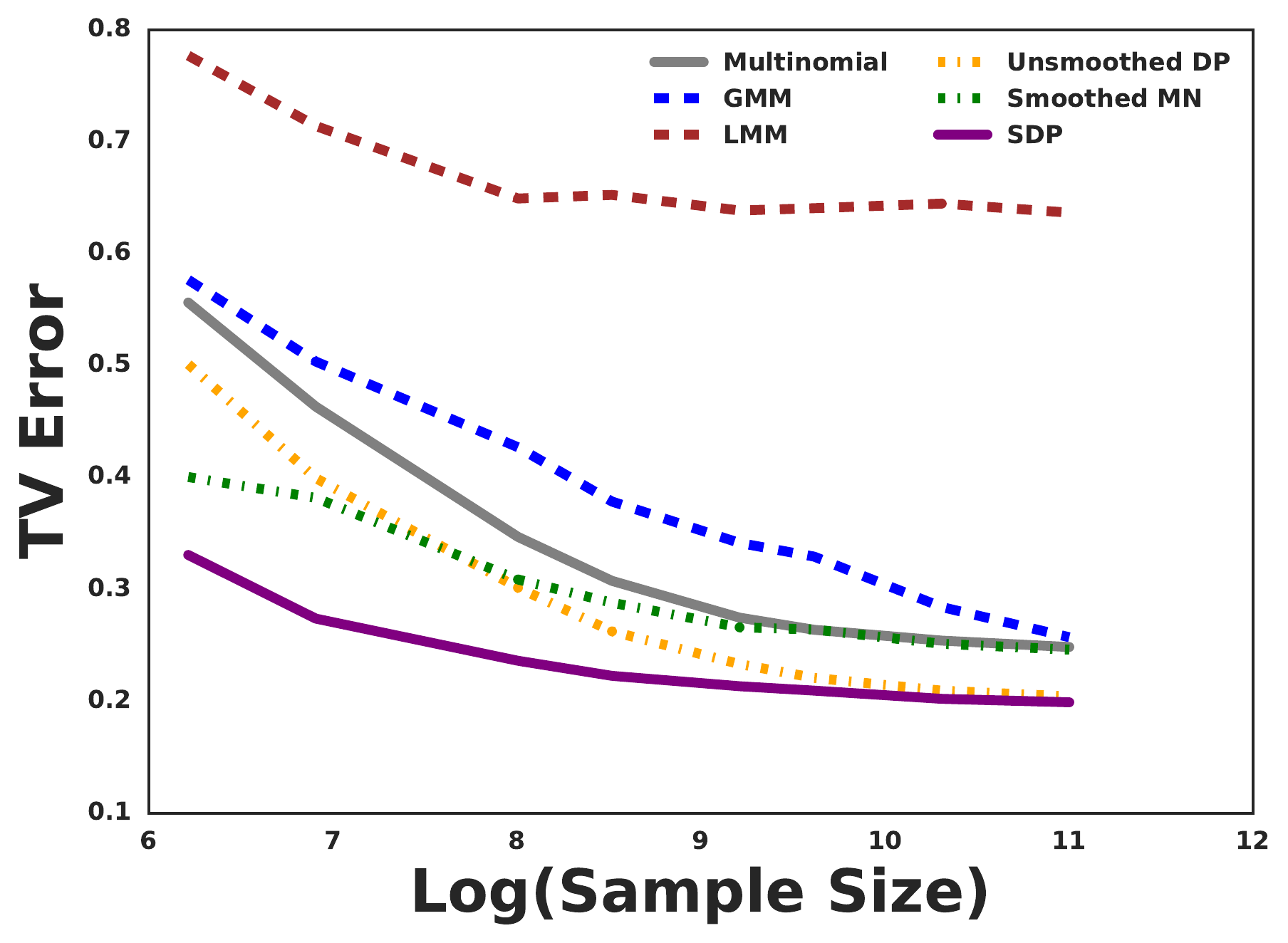}
\caption{Edge-Biased Distribution}
\end{subfigure}
\caption{\label{fig:synthetic_scores} Performance of each method on the latent GMM (left) and edge-biased (right) distributions as the sample size increases. The completely smooth GMM distribution uses a 3-component Gaussian mixture model and has no modes near the boundaries. The result is an easy task for a GMM (mixture density network) model, though the SDP model still outperforms all the other misspecified models and is competitive in the small-sample regime. The edge-biased distribution has peaks at both boundaries, similar to observed pixel intensities in natural images, and here SDP performs very well. In comparison to its constituent strategies (Unsmoothed DP and Smoothed MN), we see that SDP effectively sees an additive boost in low-sample regimes by combining the two methods.}
\end{figure*}

We compare six methods:
\begin{itemize}
    \item \textbf{Multinomial (MN)}: A simple multinomial model with no knowledge of the structure of the underlying space.
    \item \textbf{Gaussian Mixture Model (GMM)}: An $m$-component GMM or Mixture Density Network (MDN) \cite{bishop:mdn:1994}. For multi-dimensional data, we use a Cholesky parameterization of the covariance matrix.
    \item \textbf{Logistic Mixture Model (LMM)}: An $m$-component mixture of logistics, implemented using the CDF method of PixelCNN++ \cite{salimans:etal:2017:pixelcnnpp}.
    \item \textbf{Unsmoothed Dyadic Partitions (UDP)}: Our dyadic partitioning model with no smoothing.
    \item \textbf{Smoothed Multinomial (SMN)}: A multinomial model where structure of the space is added by applying a trend filtering penalty on the logits.
    \item \textbf{Smoothed Dyadic Partitions (SDP)}: Our model with dyadic partitioning and a local smoothing window.
\end{itemize}
The first three methods have all been used in recent works in the literature. The UDP and SMN models are similar to ablation models in that they evaluate the effectiveness of SDP if one component were removed.

We consider two different ground truth distribution classes, both one dimensional. The first uses a 3-component GMM where component means and standard deviations are sampled uniformly from the range $[1,7]$ and $[0.3,2]$, respectively. The model is then discretized by evaluating the PDF at an evenly-spaced (zero-indexed) 128-bin grid along the range $[0.1,10]$. The resulting distribution always has modes that fall far away from the boundaries at 0 and 127.

This does not reflect the typical nature of real discrete data, however, which often exhibits spikes near the boundaries. To address these cases, we generate a second set of experiments where the ground truth is drawn from a mixture model of the following form:
\begin{equation}
p(x) = \frac{1}{3}\text{Exp}(x | \lambda_1) + \frac{1}{3}\text{Exp}(10.1-x | \lambda_2) + \frac{1}{3}\mathcal{N}(x | \mu, \sigma) \, ,
\label{eq:edge_biased_dist}
\end{equation} 
where $\text{Exp}$ is the exponential distribution.  We sample $\lambda_1$ and $\lambda_2$ uniformly randomly from the range $[0.25,2]$ and sample $\mu$ and $\sigma$ as in the GMM, then discretize this method following the same procedure used for the GMM. This creates an \textit{edge-biased} distribution, where a smooth mode exists somewhere in the middle of the space, but at the boundaries the probability mass increases exponentially---similar to the observed marginal subpixel intensity in the CIFAR dataset \cite{salimans:etal:2017:pixelcnnpp}.

For both distributions, we evaluate all six models on sample sizes of 500, 1K, 3K, 5K, 10K, 15K, 30K, and 60K. The base network architecture for each model uses two $5\times5$ convolution layers of size 32 and 64 with $2\times2$ max pooling, followed by a dense hidden layer of size $1024$; all layers use ReLU activations and dropout. All models are trained for 100K steps using Adam with a learning rate of $10^{-4}$, $\epsilon = 1$, and batchsize of $50$ with $20\%$ of the training samples used as a validation set and validation every 100 steps used to save the best model and prevent overfitting (i.e.~the overfitting problems noted in Section \ref{subsec:experiments:neighborhoods} are not an issue here). For the GMM and LMM models, we evaluated over $m=\{1,3,5,10,20\}$. For the smoothed models, we fixed the neighborhood radius at 5 and evaluted over $k=\{1,2\}$ and $\lambda=\{0.0001,0.0005,0.001,0.005,0.01,0.05,0.1,0.5,1.0\}$. All hyperparameters were selected using the validation set performance.

\begin{figure*}[!ht]
\centering
\begin{subfigure}[t]{0.32\textwidth}
\includegraphics[width=\textwidth]{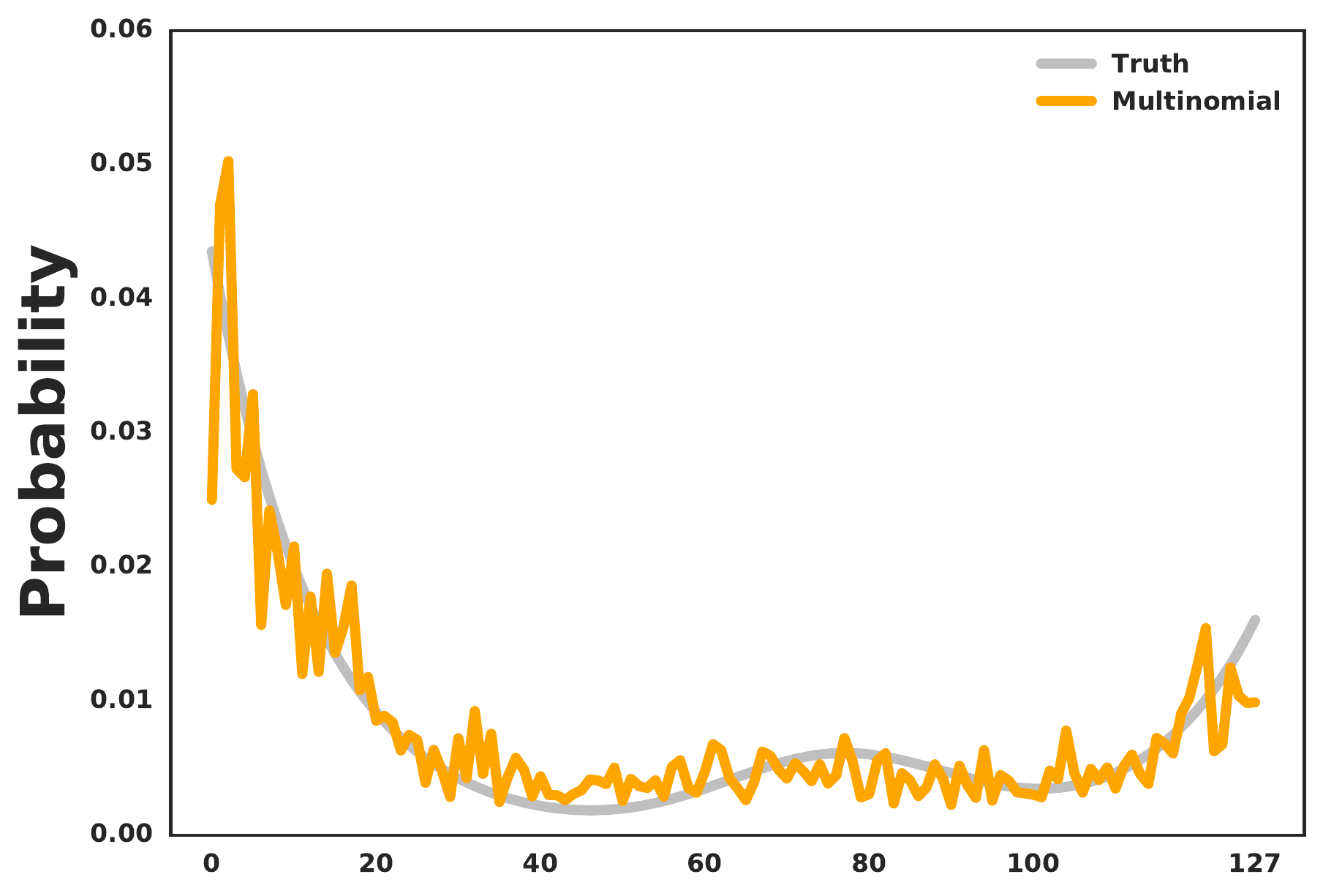}
\caption{Multinomial}
\end{subfigure}
\begin{subfigure}[t]{0.32\textwidth}
\includegraphics[width=\textwidth]{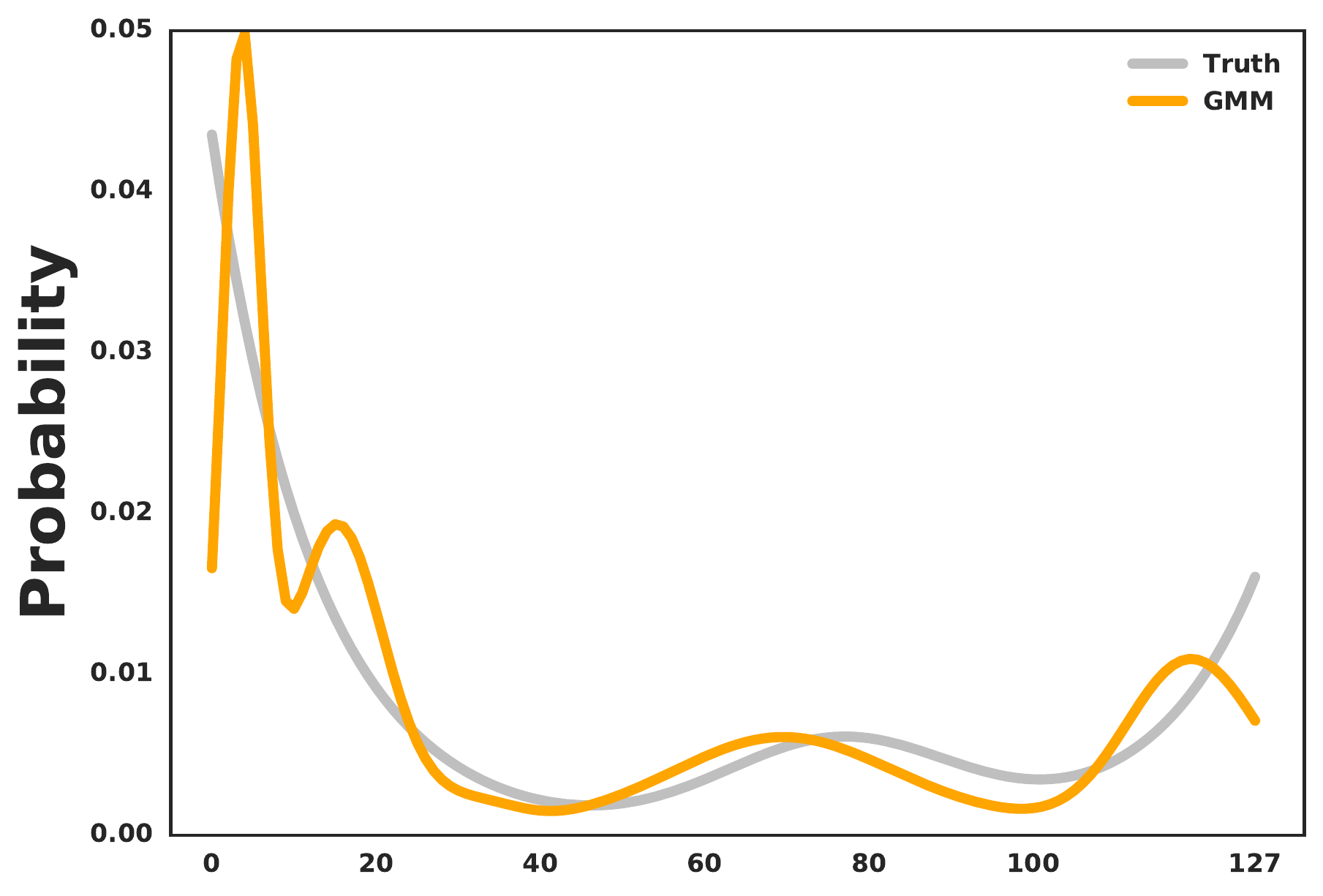}
\caption{GMM}
\end{subfigure}
\begin{subfigure}[t]{0.32\textwidth}
\includegraphics[width=\textwidth]{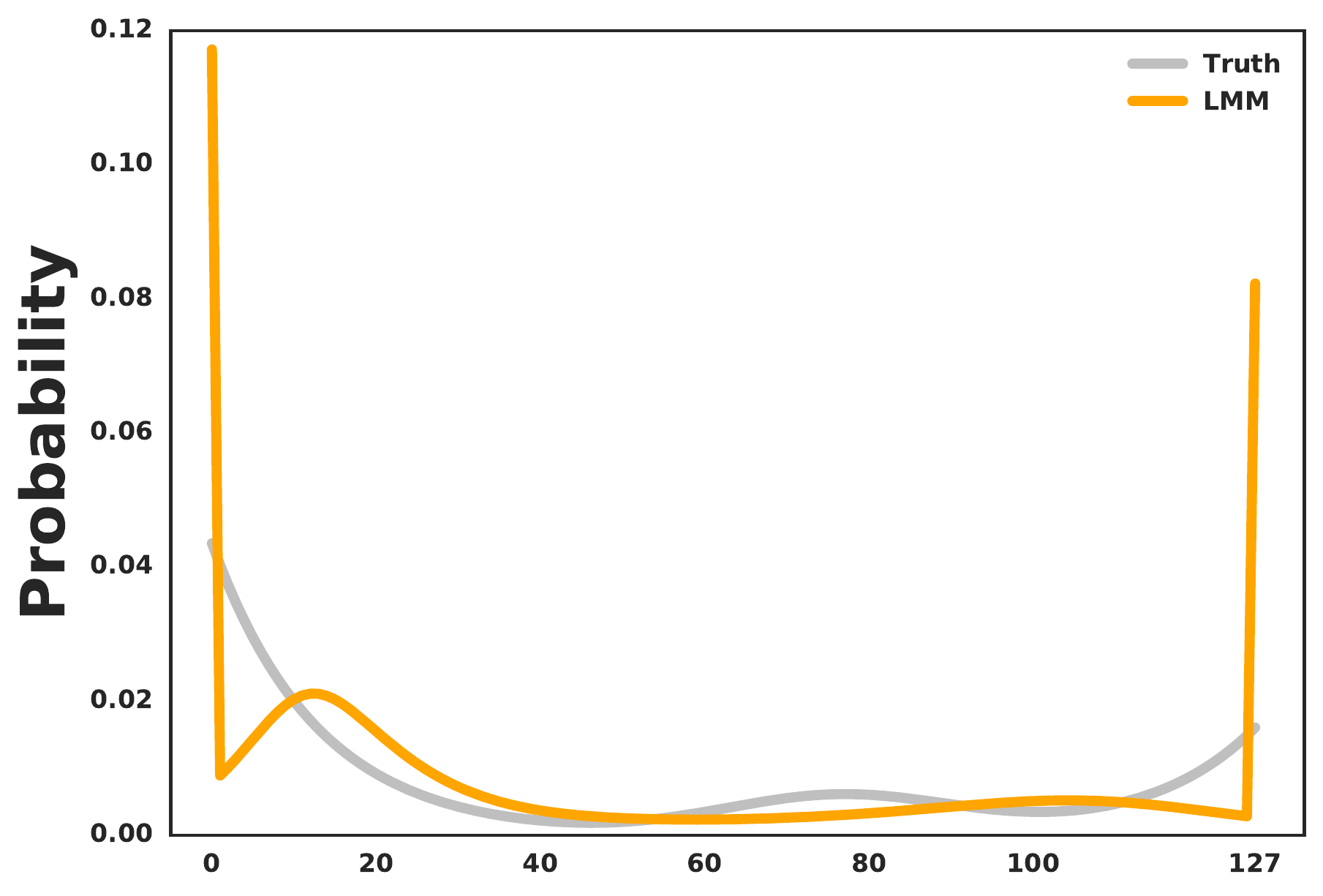}
\caption{LMM}
\end{subfigure}
\begin{subfigure}[t]{0.32\textwidth}
\includegraphics[width=\textwidth]{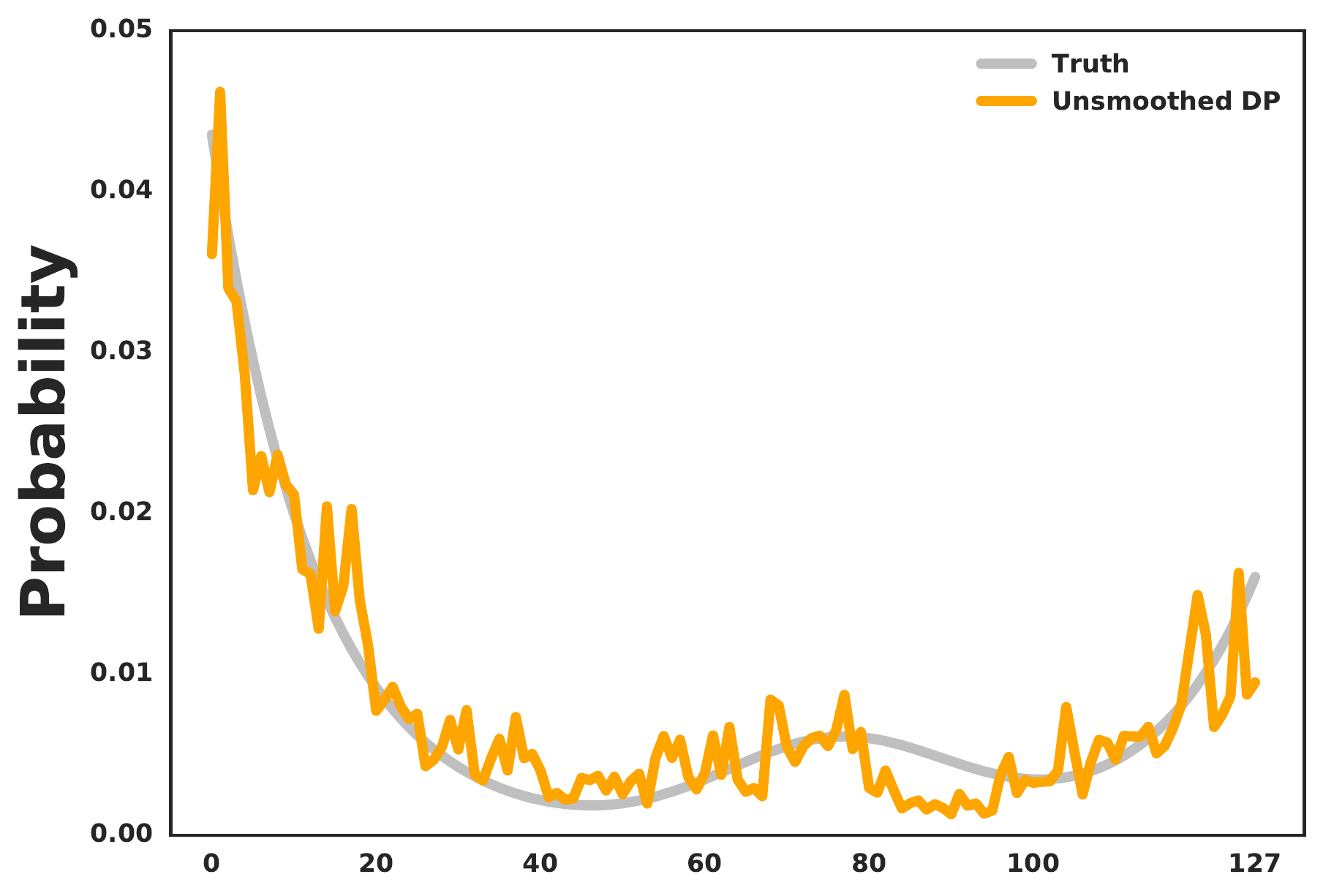}
\caption{Unsmoothed DP}
\end{subfigure}
\begin{subfigure}[t]{0.32\textwidth}
\includegraphics[width=\textwidth]{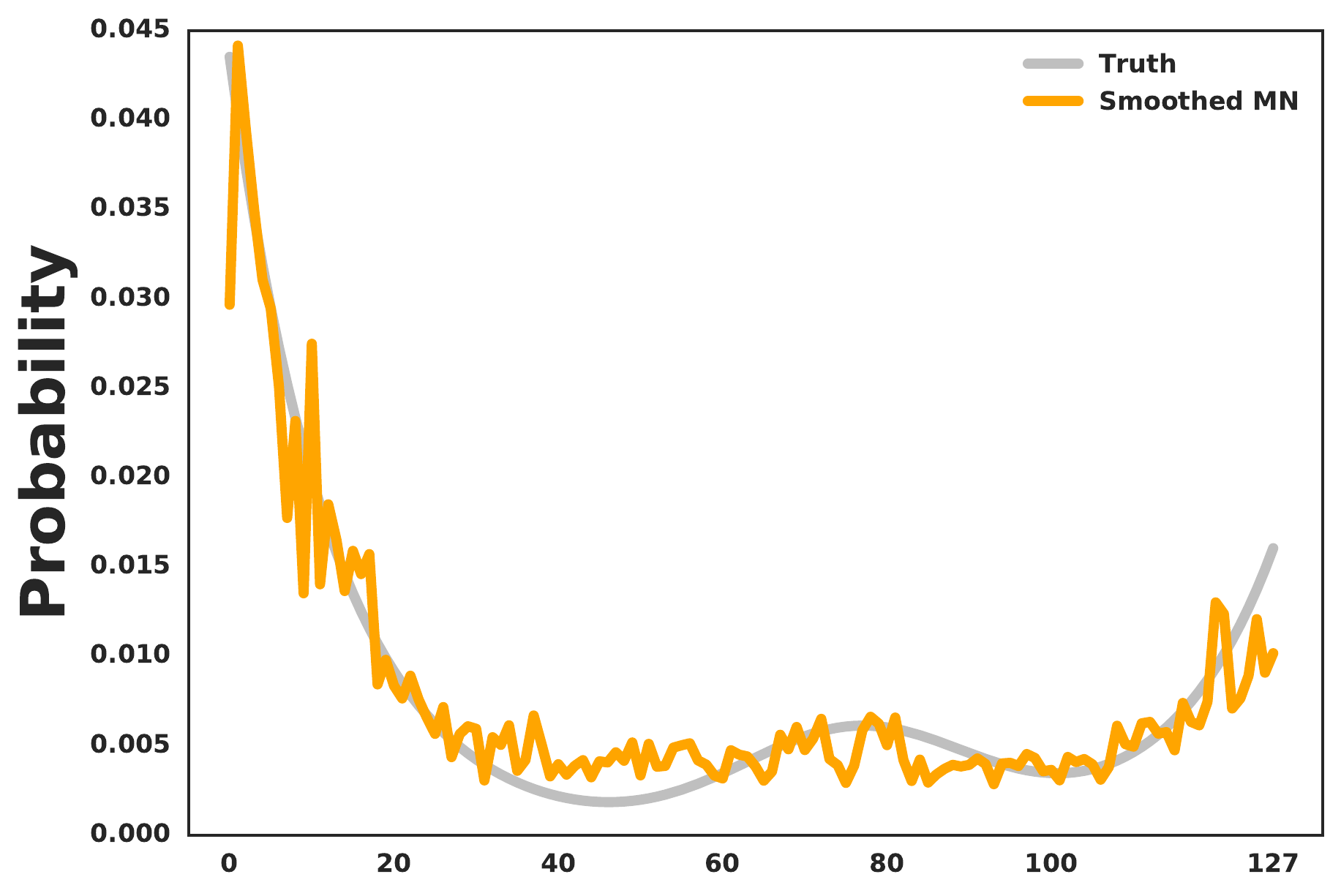}
\caption{Smoothed MN}
\end{subfigure}
\begin{subfigure}[t]{0.32\textwidth}
\includegraphics[width=\textwidth]{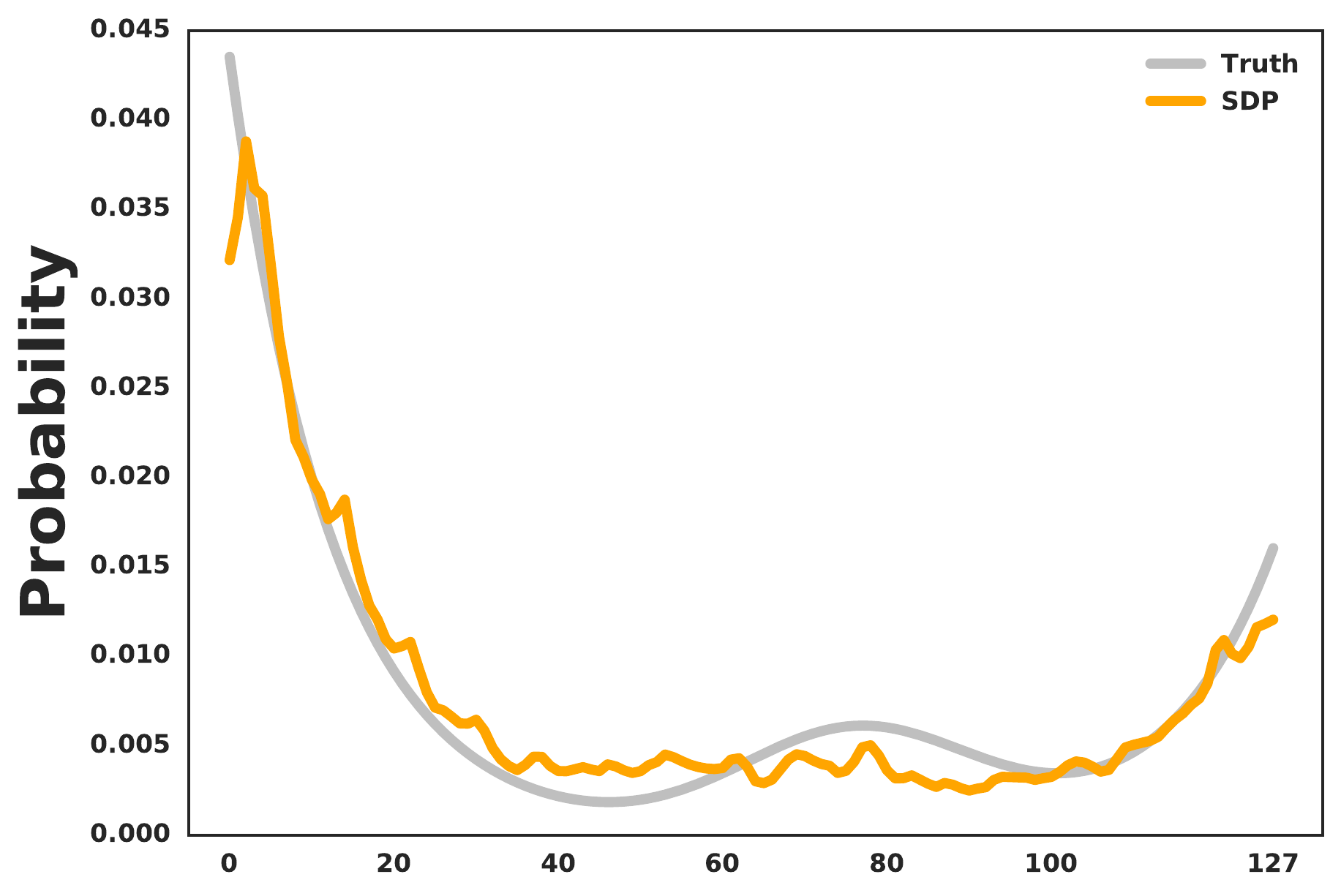}
\caption{SDP}
\end{subfigure}
\caption{\label{fig:edge_biased_examples} Qualitative examples of fits for each of the benchmark methods with 3000 training samples on the edge-biased distribution. (a) The multinomial model is extremely noisy due to no knowledge of label structure. (b) The GMM model never puts substantial mass outside the feasible range, resulting in underestimation in the tails. (c) The LMM model over-estimates the boundaries due to the CDF formulation of the log-likelihood. (d and e) The unsmoothed DP and smoothed multinomial models both improve on the pure multinomial model, but still have a high degree of noise. (f) The SDP model finds a smooth fit which does not grossly misestimate the tails.}
\end{figure*}

Figure \ref{fig:synthetic_scores} shows the results, measured in terms of total variation distance from the true distribution, averaged across ten independent trials. For the GMM distribution (Figure \ref{fig:synthetic_scores}a), the GMM model is well-specified and consequently performs very well. In the low-sample GMM regime, the SDP model is competitive with the GMM model, despite the fact that the GMM  matches the parametric form of the ground truth.  As previously noted, however, most data sets do not follow such an ideal form; for example, previous work \cite{oord:etal:2016:wavenet,oord:etal:2016:pixelcnn,oord:etal:2016:pixelrnn} has noted a multinomial model often outperforms a GMM model.  If the GMM distribution were reflective of real data, we would not expect the multinomial model to outperform it.

The edge-biased results in Figure \ref{fig:synthetic_scores}b may be of more practical interest, as the design of this experiment is directly motivated by the real marginal subpixel intensity distributions seen in natural images. In the edge-biased scenario the multinomial model does in fact outperform the GMM model. However SDP is clearly the best model here, with much stronger performance across all sample sizes. Interestingly, the LMM model performs very poorly, despite its design also being inspired by modeling pixel intensities. To better understand the performance of each of the models on the edge-biased dataset, we generated example conditional distributions when the model is trained with 3K samples.

Figure \ref{fig:edge_biased_examples} shows plots of each model's estimation of the conditional distribution of the label for a single example image, with the ground truth shown in gray. The multinomial model (Figure \ref{fig:edge_biased_examples}a) treats every value as independent and results in a jagged reconstruction, especially in the tails where the variance is particularly high. The GMM model (Figure \ref{fig:edge_biased_examples}b) provides a smooth estimation that captures the middle mode well, but it drastically underestimates the tails because of the symmetric assumption of the model components. Conversely, the LMM (Figure \ref{fig:edge_biased_examples}c) produces large spikes at the two boundaries. This is due to the formulation of the model where the boundaries are taken as the total component mass from $[-\infty,0]$ and $[127,\infty]$. This is an intentional bias in the model designed to better match CIFAR pixel intensities which also have spikes at the boundaries.  But it is quite a strong bias, as it effectively results in a two-point-inflated smooth model with a nontrivial bias towards the boundaries. Finally, the UDP and SMN models (Figures \ref{fig:edge_biased_examples}d and \ref{fig:edge_biased_examples}e) result in slightly better fits than the simple multinomial model, but the combination of the two in the SDP model (Figure \ref{fig:edge_biased_examples}f) results in a smooth fit that is able to estimate the tails well.

\begin{figure}
\centering
\includegraphics[width=0.45\textwidth]{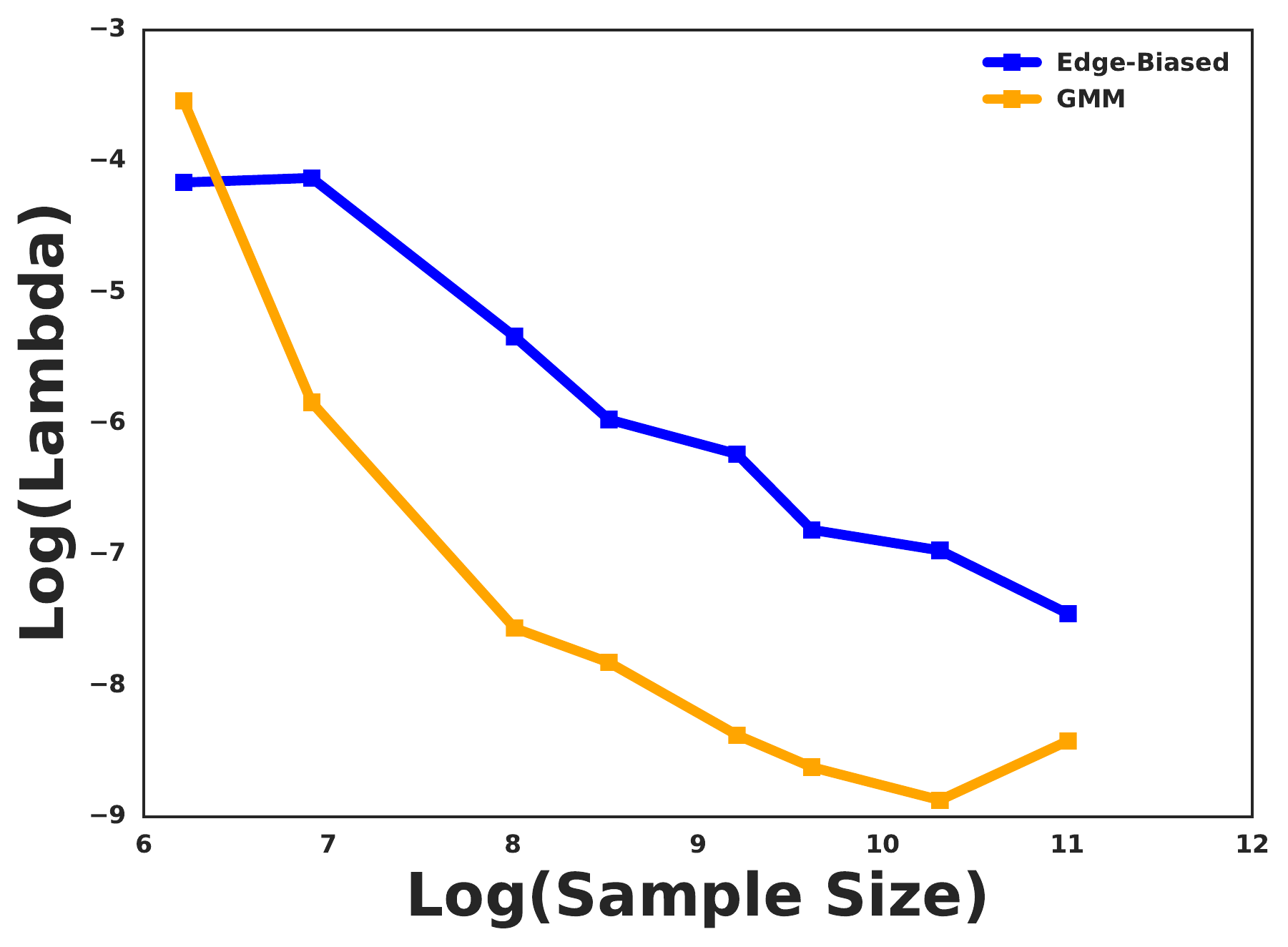}
\caption{\label{fig:synthetic_lambdas} The average selected lambda penalty parameter for the SDP model on the GMM and edge-biased distributions as a function of the sample size. In both cases, the model smooths progressively less as the sample size increases and eventually converges to the unsmoothed model in large-sample regimes.}
\end{figure}

In both distributions, we observe that the dyadic partitioning and local smoothing are jointly beneficial. Both unsmoothed DP and  smoothed MN outperform a simple multinomial, but combining them both in the SDP model is superior to both. As the sample size grows, the SDP model converges to the UDP model in performance. This is unsurprising, as increased data results in a decreased advantage to smoothing. Indeed, as we show in Figure \ref{fig:synthetic_lambdas}, the average chosen $\lambda$ value (i.e.~the amount of smoothing) decreases as the sample size grows.

\subsection{Real-world datasets}
\label{subsec:experiments:realworld}

As a final validation of our method, we compile a benchmark of real-world datasets with discrete conditional distributions as a target output (or where the target variable is discrete). We use seven datasets from the UCI database; three are one-dimensional targets, three are two-dimensional, and one is three-dimensional. Every model uses a network architecture of three hidden layers of sizes 256, 128, and 64 with ReLU activation, weight decay, and dropout. All models were trained with Adam and a decaying learning rate with initial rate at $10^{-1}$, minimum rate at $10^{-4}$, and decay rate of $0.25$; we dynamically schedule the decay by decaying the rate after the current model has failed to improve for 10 epochs. Training stops after 1000 epochs or if the current learning rate is below the minimum learning rate or training. All results are averages using 10-fold cross-validation and we use 20\% of the training data in each trial as a validation set. For all datasets, we select hyperparameter settings as in Section \ref{subsec:experiments:synthetic}. We plot the marginal distributions of each real dataset in Figure \ref{fig:marginals}.

We also evaluate on a pixel prediction task for both MNIST and CIFAR-10, where we sample a $10\times10$ patch of the image and must predict the pixel located at $(11, 11)$, relative to the origin of the patch. For both image datasets, we consider 3 different training sample sizes (500, 5K, and 50K). Every model uses a network architecture of two $3\times3$ convolution layers of size 32 and 64, with $2\times2$ max pooling, followed by three dense hidden layers of size 1024, 128, and 32; all layers use ReLU activation, weight decay, and dropout. Other training details are identical to the UCI dataset, with the exception that we only perform a single trial on the CIFAR datasets due to computational constraints. Similarly, we reduce the resolution of the CIFAR dataset from $256^3$ to $64^3$. Plots of the marginal distributions of all our datasets are available in the supplementary material.

\begin{table*}
\centering
\begin{small}
\begin{tabular}{|l|l|l|ll|ll|ll|ll|}
\hline
\multicolumn{1}{|c}{} & \multicolumn{1}{|c|}{} & \multicolumn{1}{|c|}{} & \multicolumn{2}{|c|}{Multinomial} & \multicolumn{2}{|c|}{GMM} & \multicolumn{2}{|c|}{LMM} & \multicolumn{2}{|c|}{SDP} \\ \hline
Model & Grid Size & Samples & log-probs & RMSE & log-probs & RMSE & log-probs & RMSE & log-probs & RMSE \\ \hline

Abalone  & 29 & 4177 & \textbf{-822.83} &  \textbf{2.17} & -907.78 &  2.42 & -857.23 &  2.30 & -851.88 &  2.31 \\
Auto-MPG & 377 & 392 & -177.81 & 37.34 & -187.02 & 31.59 & -186.67 & 32.49 & \textbf{-160.31} & \textbf{30.84} \\ 
Housing  & 451 & 506 & -297.59 & 69.20 & -247.23 & 40.81 & \textbf{-240.20} & 39.53 & -246.44 & \textbf{36.18} \\ 
MNIST-500  & 256 & 500 & \textbf{-1416.69} & \textbf{80.90} & -2588.79 &  92.78 & -1658.78 &  81.57 & -1466.65 &  86.68 \\
MNIST-5K  & 256 & 5000 & -1229.77 &  64.10 & -2096.24 &  94.16 & -1231.01 &  64.29 & \textbf{-1224.42} & \textbf{63.06} \\
MNIST-50K  & 256 & 50000 & -1173.80 &  58.82 & -2365.57 &  94.24 & -1191.11 &  60.41 & \textbf{-1161.69} & \textbf{57.16} \\ \hline

Students  & $21\times20$ & 395 & -209.07 &  5.27 & -219.67 &  5.44 & -209.43 &  5.26 & \textbf{-200.76} & \textbf{5.18} \\
Energy  & $38\times38$ & 768 & -323.10 &  6.21 & -492.49 &  10.89 & -437.90 &  14.24 & \textbf{-279.01} &  \textbf{4.17} \\
Parkinsons  & $36\times49$ & 5875 & \textbf{-1941.91} & \textbf{6.42} & -3969.63 &  10.91 & -3633.29 &  13.53 & -3530.22 &  14.93 \\ \hline

Concrete  & $30\times59\times43$ & 103 & -115.88 &  21.34 & -107.46 &  18.91 & -108.63 &  21.07 & \textbf{-102.34} & \textbf{18.09} \\
CIFAR-500  & $64\times64\times64$ & 500 & -9980.57 &  26.35 & -9177.59  & \textbf{24.69} & -9109.57 & 26.00 & \textbf{-8519.21} & 25.81 \\
CIFAR-5K  & $64\times64\times64$ & 5000 & -9688.08 &  26.26 & -9106.04 & 23.01 & -9213.49 &  26.11 & \textbf{-7504.35} & \textbf{14.89} \\
CIFAR-50K  & $64\times64\times64$ & 50000 & -8409.60 &  22.66 & -9099.49 & 23.02 & -9214.51 &  26.08 & \textbf{-6796.39} & \textbf{13.42} \\ \hline
\end{tabular}
\end{small}
\caption{\label{tab:realdata_results} Results for the four models on a series of discrete datasets from the UCI database and the MNIST and CIFAR-10 datasets. The best scores for each metric and dataset are bolded; grid size corresponds to the number of bins in the underlying discrete space. Overall, the SDP model performs very strongly especially in the cases where the discrete space is much larger than the sample size.}
\end{table*}

Table $\ref{tab:realdata_results}$ presents the results on all the candidate datasets, with the best-performing score in bold for each dataset and metric. We measure performance both in log-probability of the specific observed point and root mean squared error (RMSE), as the discrete space has a natural measurement of distance. In general, the SDP model performs very well in cases where the size of the discrete space dominates the sample size. The datasets where this is not the case (Abalone and Parkinsons), the multinomial model has sufficient data to model the space well. The LMM model outperforms in terms of log-probs on the Housing dataset likely due to the large peaks at the boundaries in the data (Figure \ref{fig:marginals}c in the supplement). The CIFAR dataset also has substantial peaks-- specifically at the corners-- resulting in the LMM outperforming the multinomial model as has been demonstrated previously in PixelCNN++ \cite{salimans:etal:2017:pixelcnnpp}. However, the additional structure in the dataset is much better modeled via the SDP model, which has nearly half the RMSE of the other methods.

\section{Discussion}
\label{sec:discussion}

As our experiments demonstrated, the SDP model outperforms several alternative models commonly used in the literature. In the one-dimensional case, many other models have been proposed ranging from more flexible parametric component distributions for mixture models \cite{carreau:bengio:2009} to quantile regression \cite{taylor:2000,lee:yang:2006}. Extending these models to higher dimensions is non-trivial, making them unsuitable for use in many of our target applications. Furthermore, even in the 1d case, it is often unclear \textit{a priori} which parametric components should be included in a mixture model and simply adding a large number may result in overfitting. A quantile regression model would also suffer from the same overfitting issues as the unsmoothed DP model in Section \ref{subsec:experiments:neighborhoods}, as it does not impose any smoothness explicitly. From a computational perspective, quantile regression would also require all nodes to be calculated at every iteration and would not scale well to large (finely discretized) 1d spaces.

There have also been other multidimensional models, notably the line of work in neural autoregressive models such as NADE \cite{uria:etal:2016}, RNADE \cite{uria:etal:2013}, and MADE \cite{germain:etal:2015}; and variational autoencoders \cite{kingma:etal:2013} such as DRAW \cite{gregor:etal:2015}. We see such models as complementary approaches rather than competitive approaches to SDP. For instance, one could modify the outputs of MADE to be a separate discrete distribution for each dimension rather than a single likelihood. This would also address the main scalability issue of our model. Currently SDP requires $\mathcal{O}(n)$ output nodes for a space of $n$ possible values. In the low-dimensional problems explored in this paper this was not a problem, but it quickly exceeds the memory of a GPU once one moves beyond three or four dimensions.

Our choice of local trend filtering for smoothing introduces three new hyperparameters: neighborhood radius size ($r$), the order of trend filtering ($k$), and the penalty weight ($\lambda$). The model appears to function well with a fairly small neighborhood of about $5\%$ of the underlying space, and even with just a fixed choice of a radius of 5 we performed well across all our real-world datasets. The choice of $k$ is also less important, as both linear and quadratic trend filtering tend to drastically improve on the model. Computationally, the $k=1$ penalty matrix is somewhat more efficient to use, though we have only optimized our 1d model for GPUs. The main parameter of interest is the choice of $\lambda$. In all of our experiments we chose to enumerate each value and fix it for the entire training procedure. In practice, one may wish to anneal $\lambda$ up as training proceeds so that the model can quickly converge to a jagged-but-good solution and then focus on smoothing.

Given the computational overhead of smoothing during training (around 3-4 times the wall-clock time of the other models in our experiments), one may be tempted to smooth the logits post-training, but there are several problems with such an approach. First, it would be unclear what inferential principle to use to determine the best fit. Second, given some ad-hoc approach to choosing $\lambda$, smoothing the logits post-training would impose a separate degree of smoothness to each sample conditional distribution. By using a single lambda for all training examples, we are in effect imposing the same degree of smoothness on all distributions in a way that constrains the degrees of freedom of the model explicitly across all classes. Evaluation of such a model on test data would also require generating the region (full or local) to smooth and then solving the full solution path each time, substantially impacting performance.

Finally, our experiments have all been done on relatively small problems with fairly simple models. Computational constraints prevented us from running extensive experiments with more sophisticated methods like PixelCNN++, which requires a multi-GPU machine and multiple days of training for a single model. It is our hope that other researchers with access to more substantial resources will evaluate SDP as an option in their larger-scale systems in the future.

\section{Conclusion}
\label{sec:conclusion}

We have presented SDP, a method for deep conditional estimation of discrete probability distributions. By dividing the discrete space into a series of half-spaces, SDP transforms the distribution estimation task into a hierarchical classification task which overcomes many of the disadvantages of simple multinomial modeling. These dyadic partitions are then smoothed using graph-based trend filtering on the resulting logit probabilities in a local region around the target label at training time. The combination of dyadic partitioning and logit smoothing was shown to have an additive effect in total variation error reduction in synthetic datasets. The benchmark results on both real and synthetic datasets suggest that SDP is a powerful new method that may substantially improve the performance of generative models for audio, images, video, and other discrete data.

\begin{small}
\bibliographystyle{icml2016}
\bibliography{main}
\end{small}

\appendix

\title{Smoothed Dyadic Partitioning - Appendix}
\date{} 

\maketitle

\section{Marginal distributions of real-world datasets}

Below are the marginal distributions of the real-world datasets used for benchmarking in Section \ref{subsec:experiments:realworld}. All of the datasets exhibit some degree of spatial structure, making this a difficult task for a multinomial model without a large sample size. Several of the datasets also contain spikes along boundaries that present problems for Gaussian mixture models. For example, the Housing data contains a large spike at the largest value; MNIST contains a large spike at its smallest value; the Student Performance data contains a ridge line along the upper boundary (i.e. students getting zero on one of the two tests); the Concrete data contains several high-probability points along the top of the (X,Z) and (Y,Z) border; and the CIFAR data contains spikes at the upper left and lower right corners of each 2d view.

\begin{figure}
\centering
\includegraphics[width=0.5\textwidth]{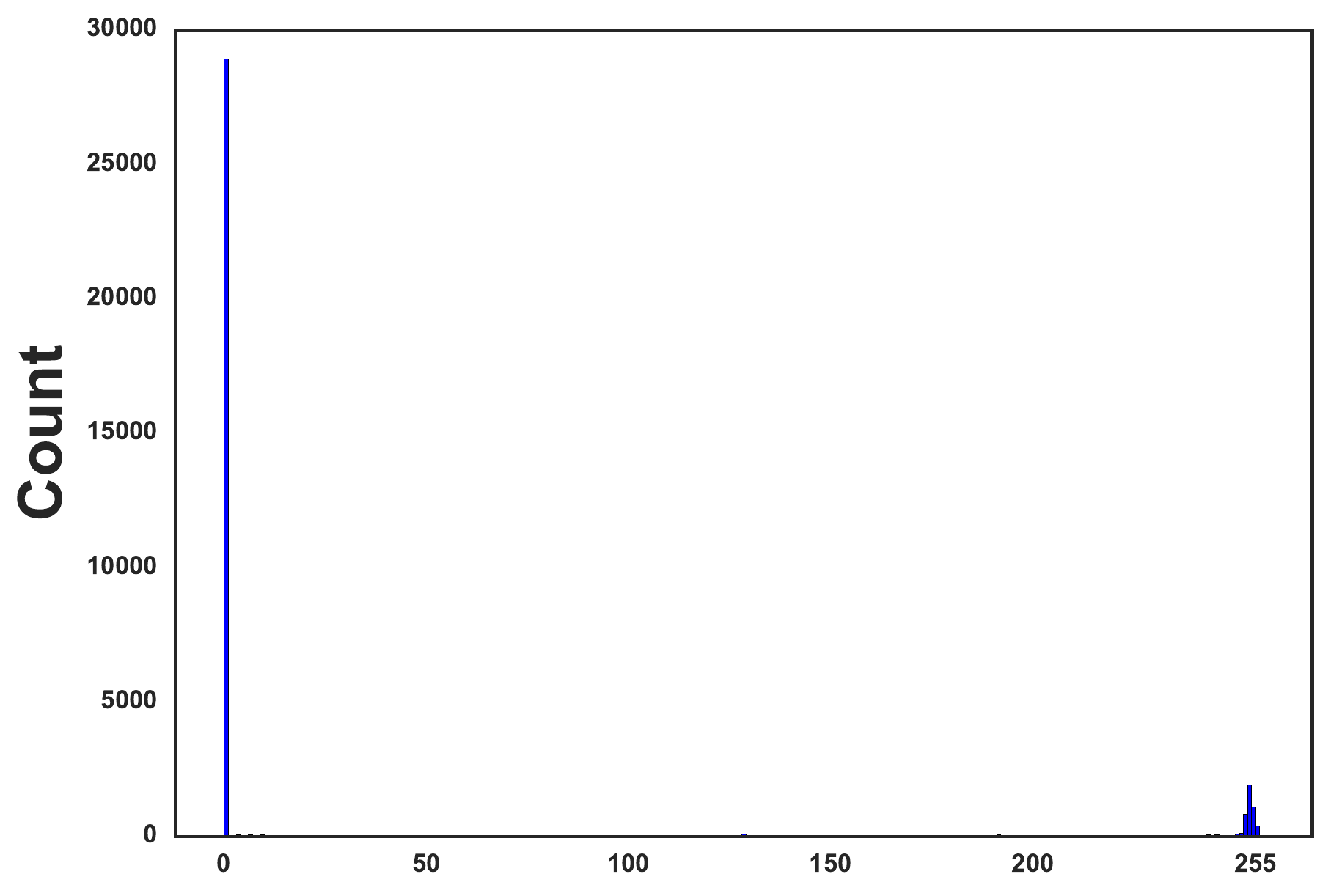}
\caption{\label{fig:marginals_mnist} The marginal pixel intensities for MNIST.}
\end{figure}

\begin{figure*}[!ht]
\centering
\begin{subfigure}[t]{0.32\textwidth}
\includegraphics[width=\textwidth]{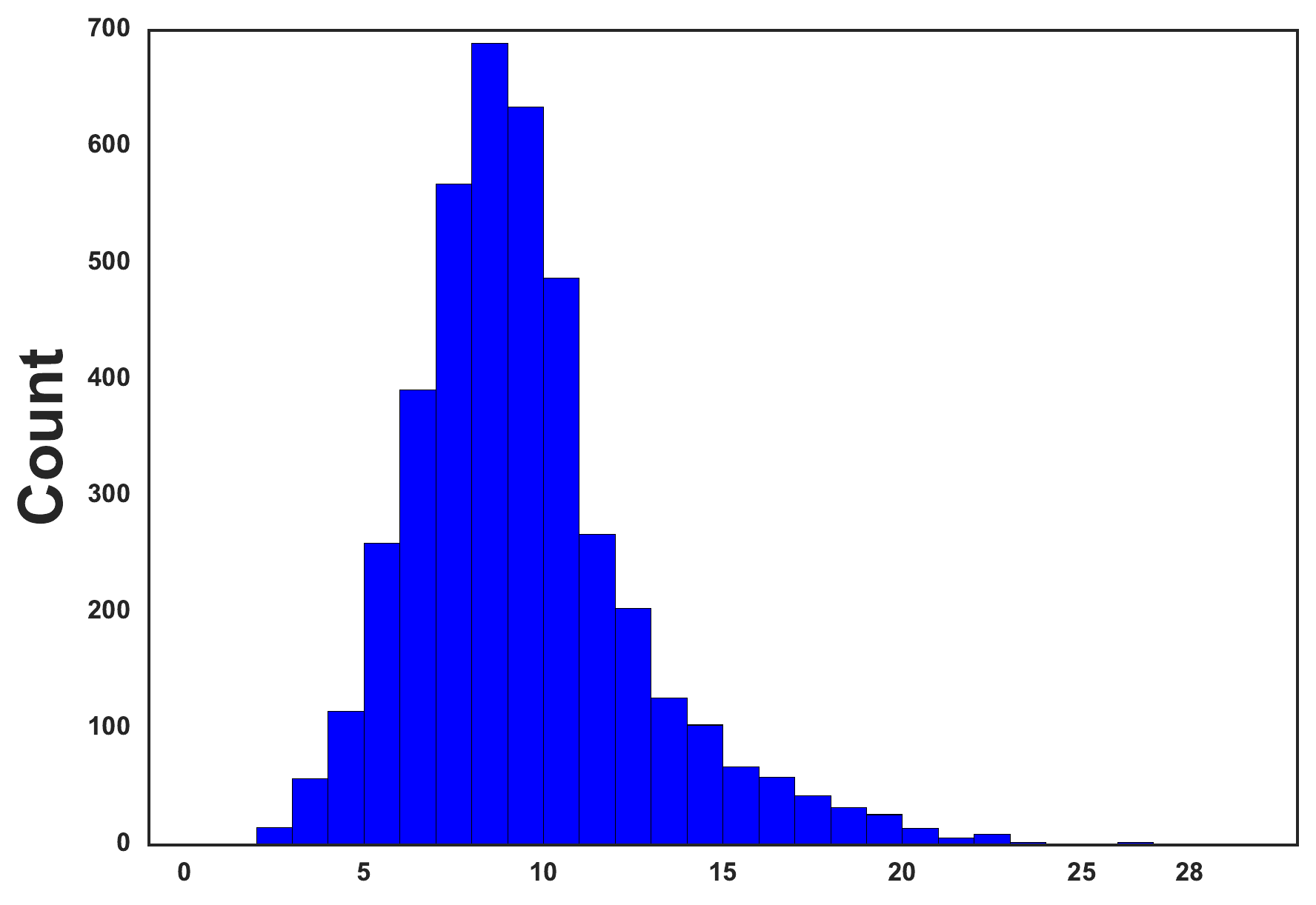}
\caption{Abalone}
\end{subfigure}
\begin{subfigure}[t]{0.32\textwidth}
\includegraphics[width=\textwidth]{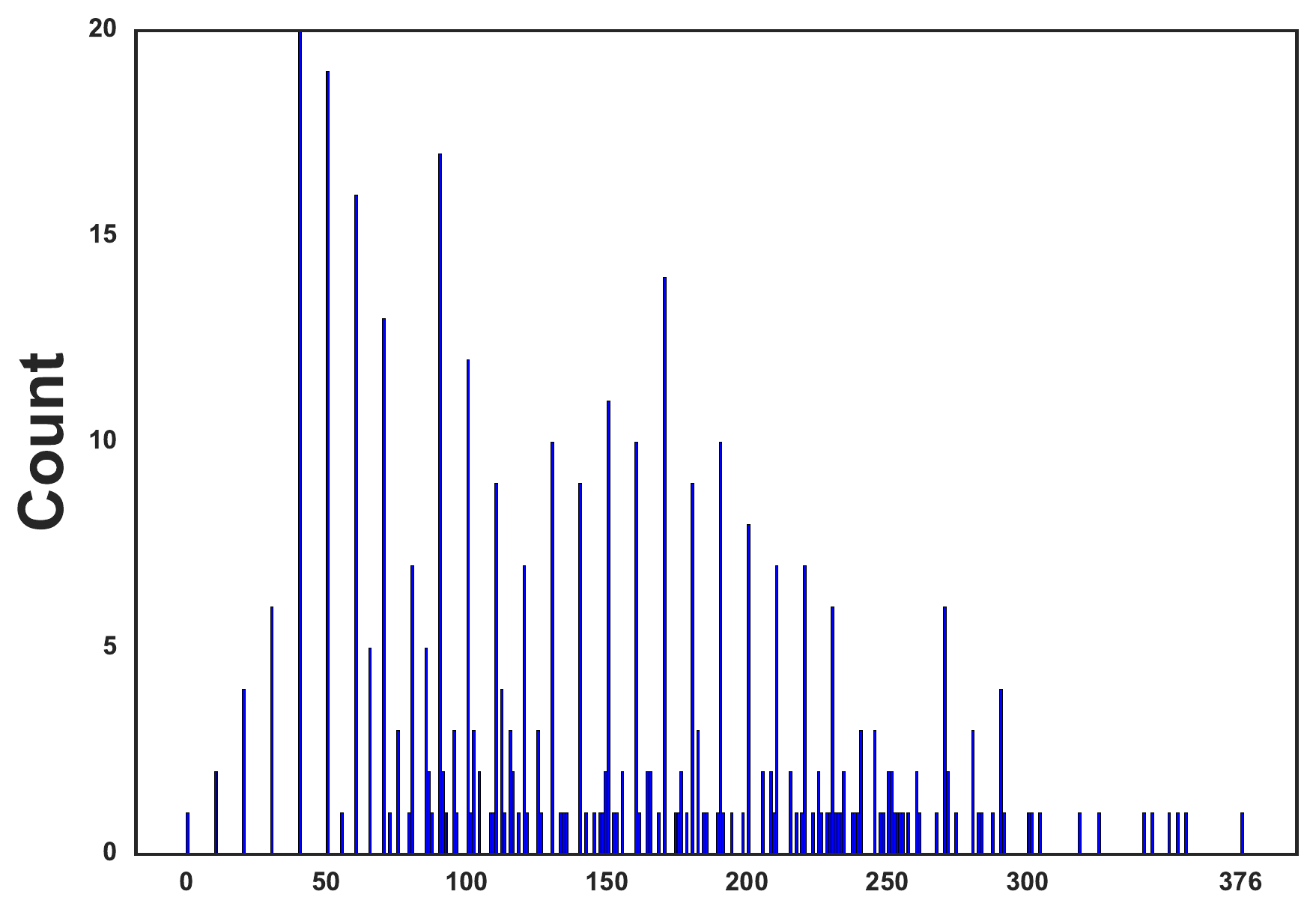}
\caption{Auto-MPG}
\end{subfigure}
\begin{subfigure}[t]{0.32\textwidth}
\includegraphics[width=\textwidth]{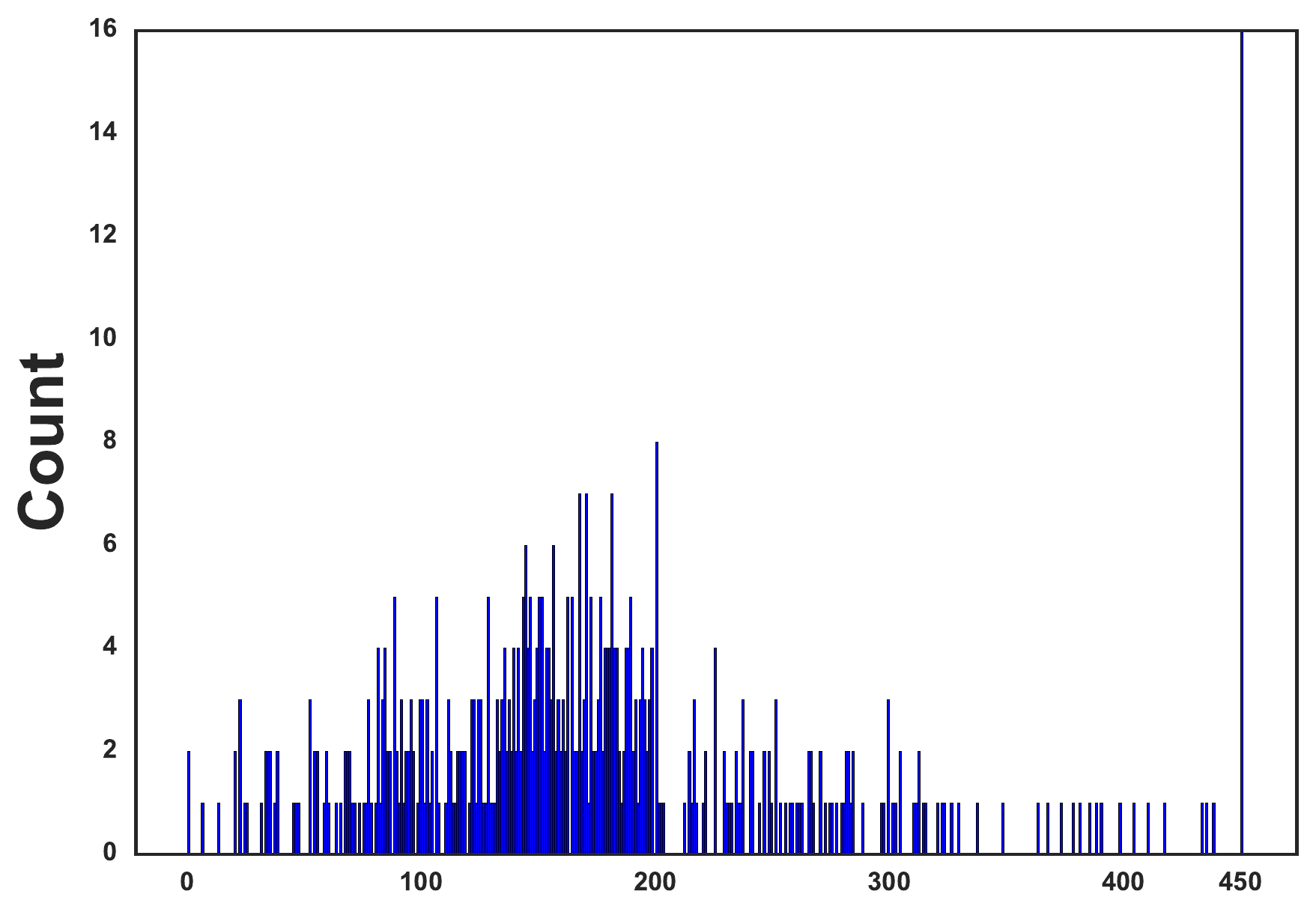}
\caption{Housing}
\end{subfigure}
\begin{subfigure}[t]{0.32\textwidth}
\includegraphics[width=\textwidth,height=5cm]{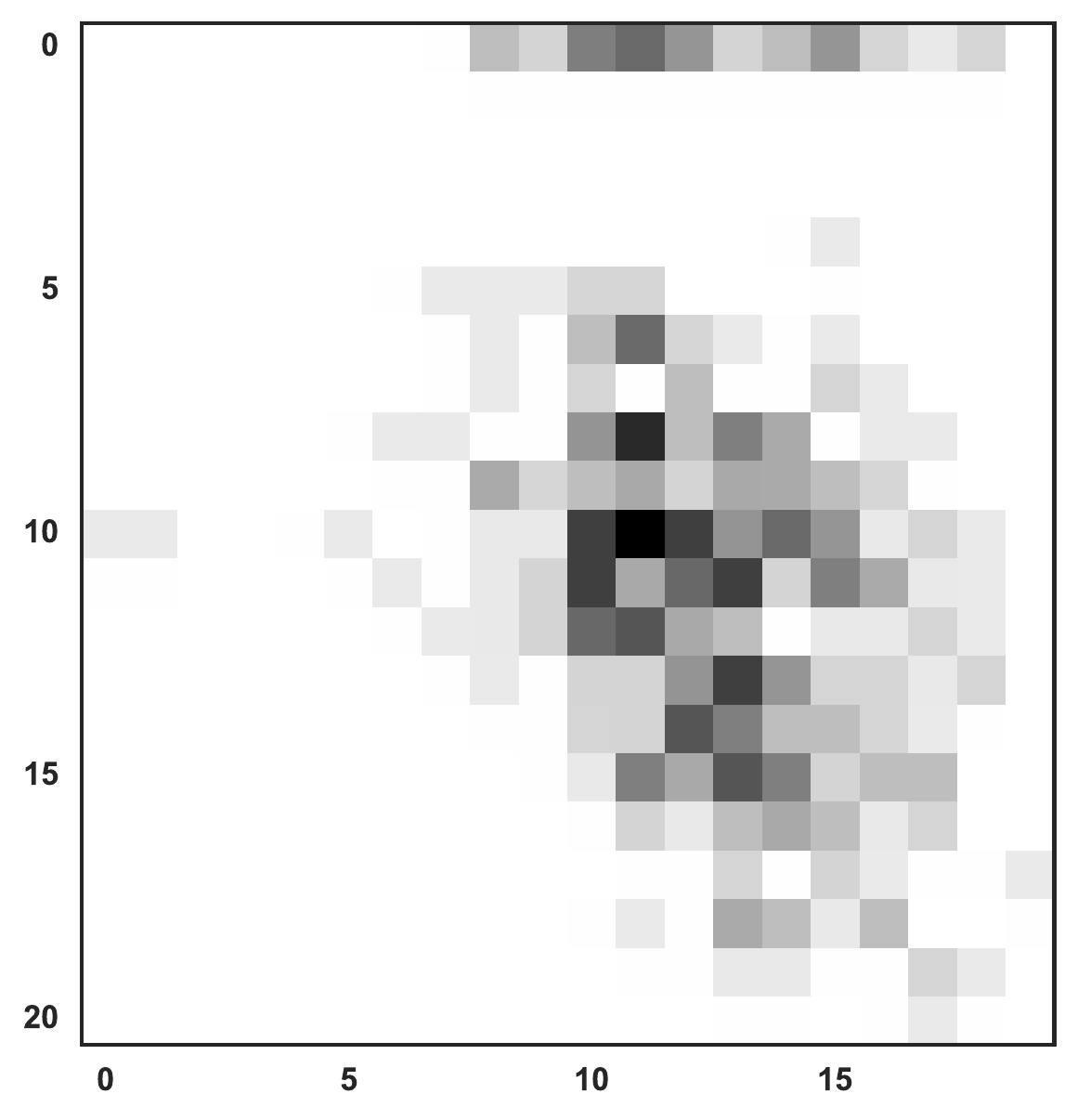}
\caption{Student Performance}
\end{subfigure}
\begin{subfigure}[t]{0.32\textwidth}
\includegraphics[width=\textwidth,height=5cm]{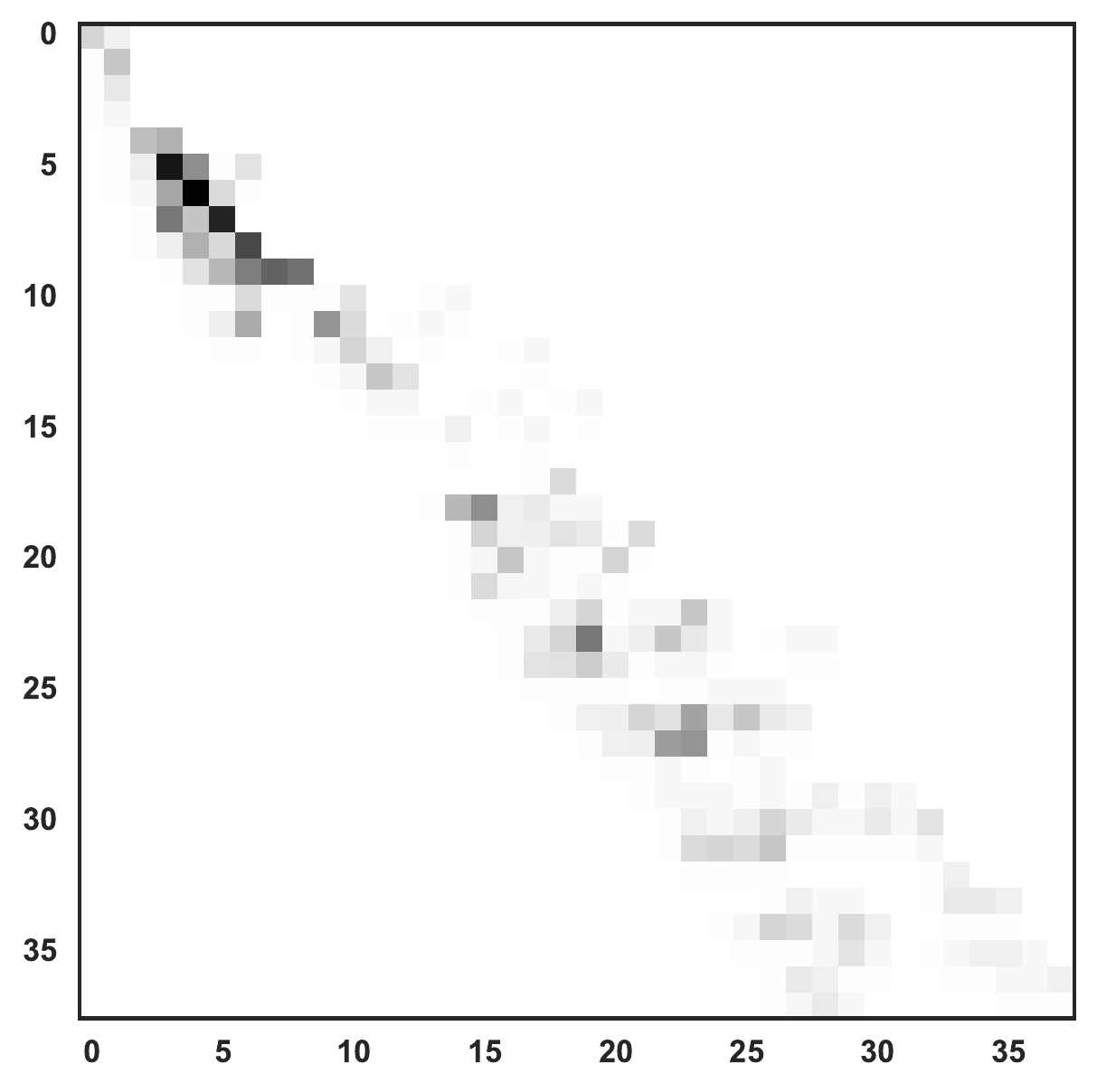}
\caption{Energy Efficiency}
\end{subfigure}
\begin{subfigure}[t]{0.32\textwidth}
\includegraphics[width=\textwidth,height=5cm]{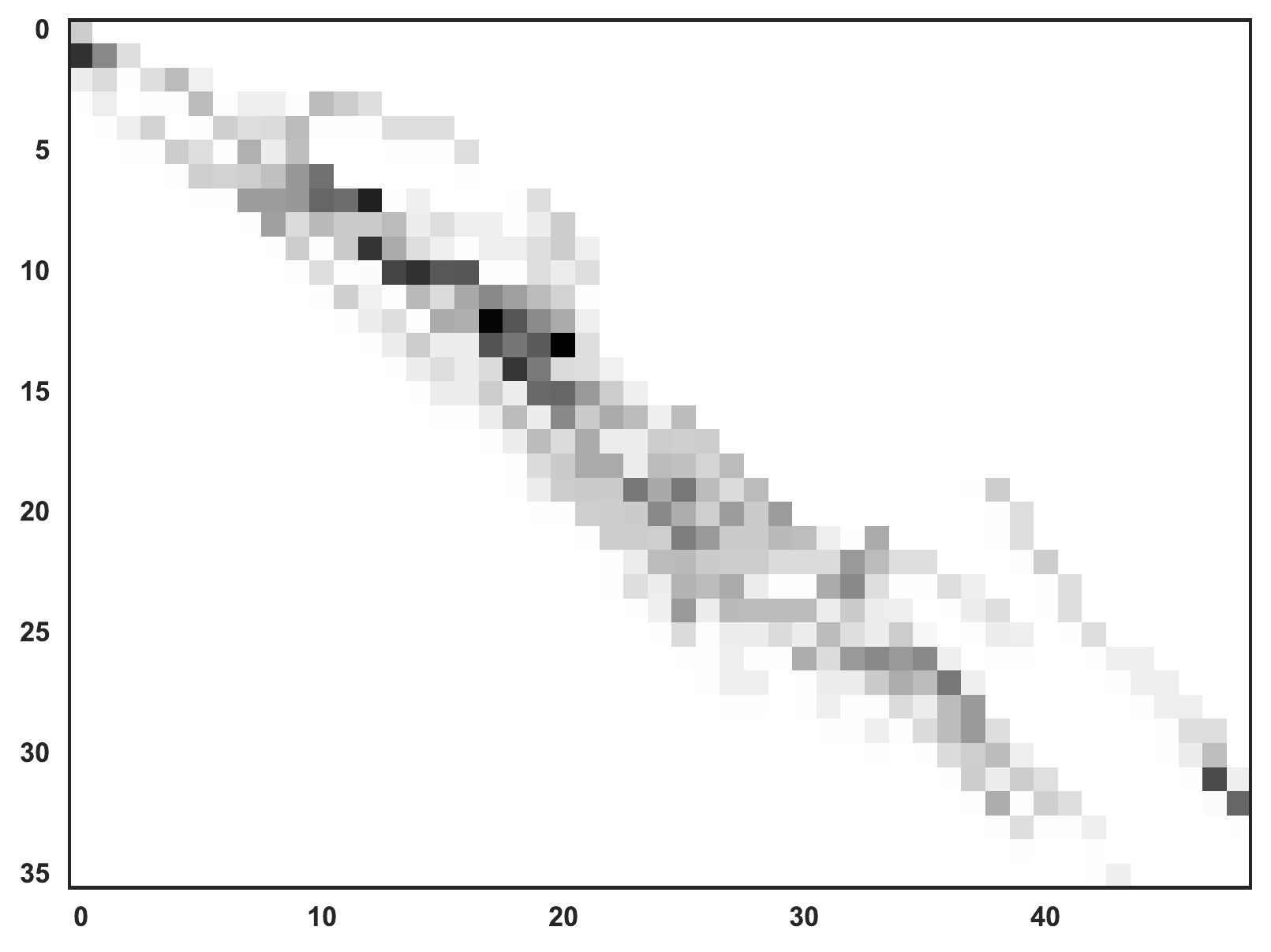}
\caption{Parkinsons Telemonitoring}
\end{subfigure}
\begin{subfigure}[t]{0.32\textwidth}
\includegraphics[width=\textwidth,height=5cm]{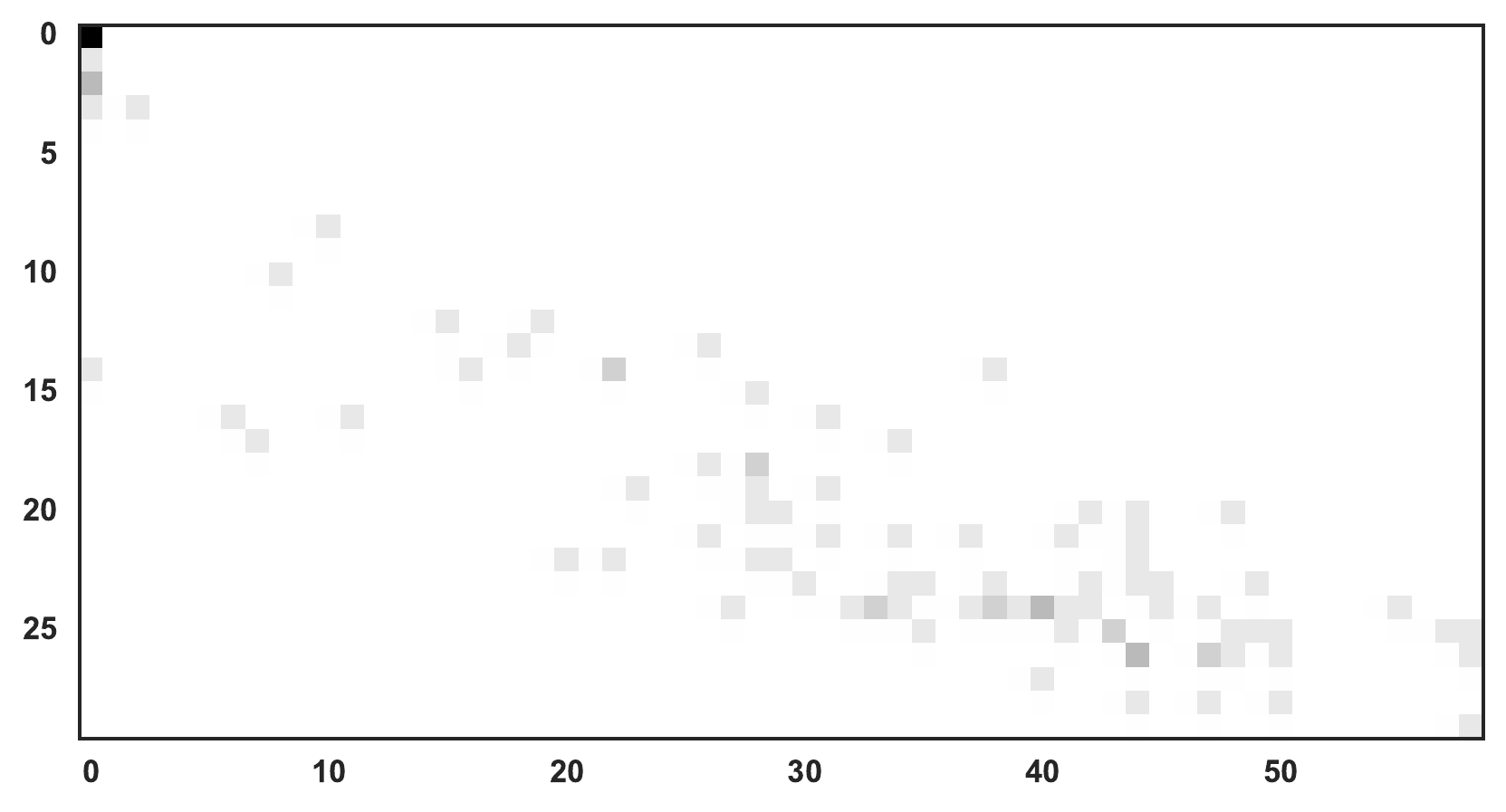}
\caption{Concrete (X,Y)}
\end{subfigure}
\begin{subfigure}[t]{0.32\textwidth}
\includegraphics[width=\textwidth,height=5cm]{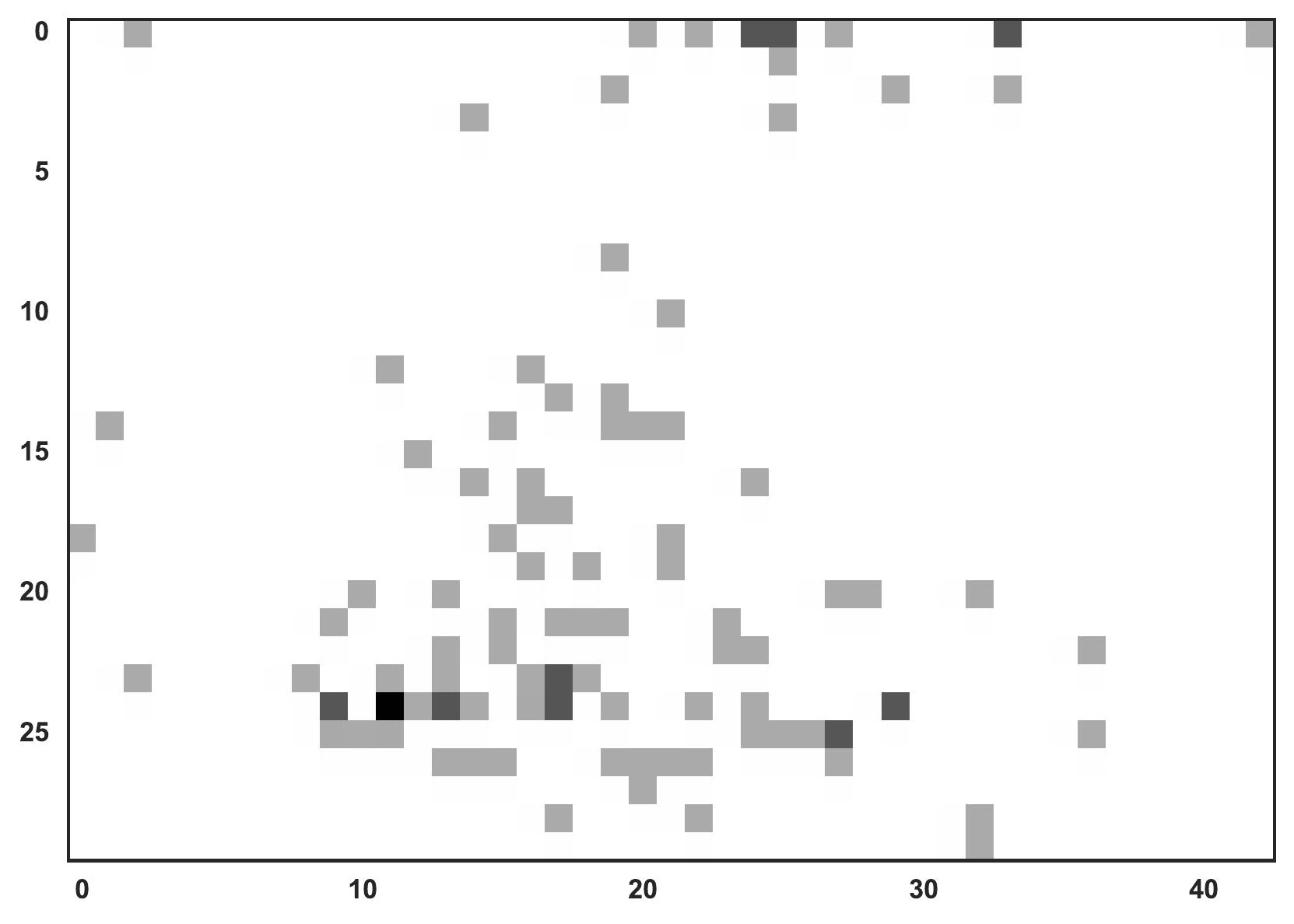}
\caption{Concrete (X,Z)}
\end{subfigure}
\begin{subfigure}[t]{0.32\textwidth}
\includegraphics[width=\textwidth,height=5cm]{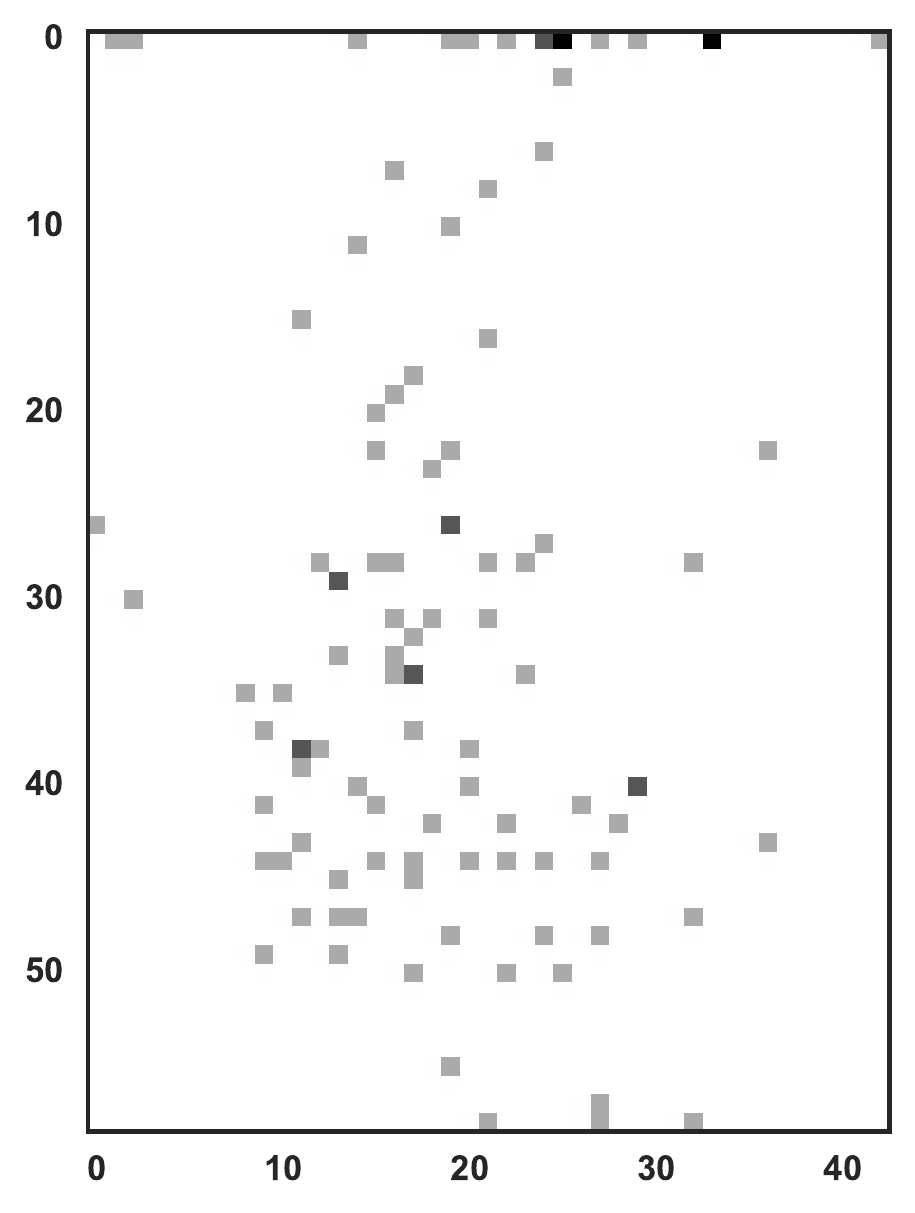}
\caption{Concrete (Y,Z)}
\end{subfigure}
\begin{subfigure}[t]{0.32\textwidth}
\includegraphics[width=\textwidth]{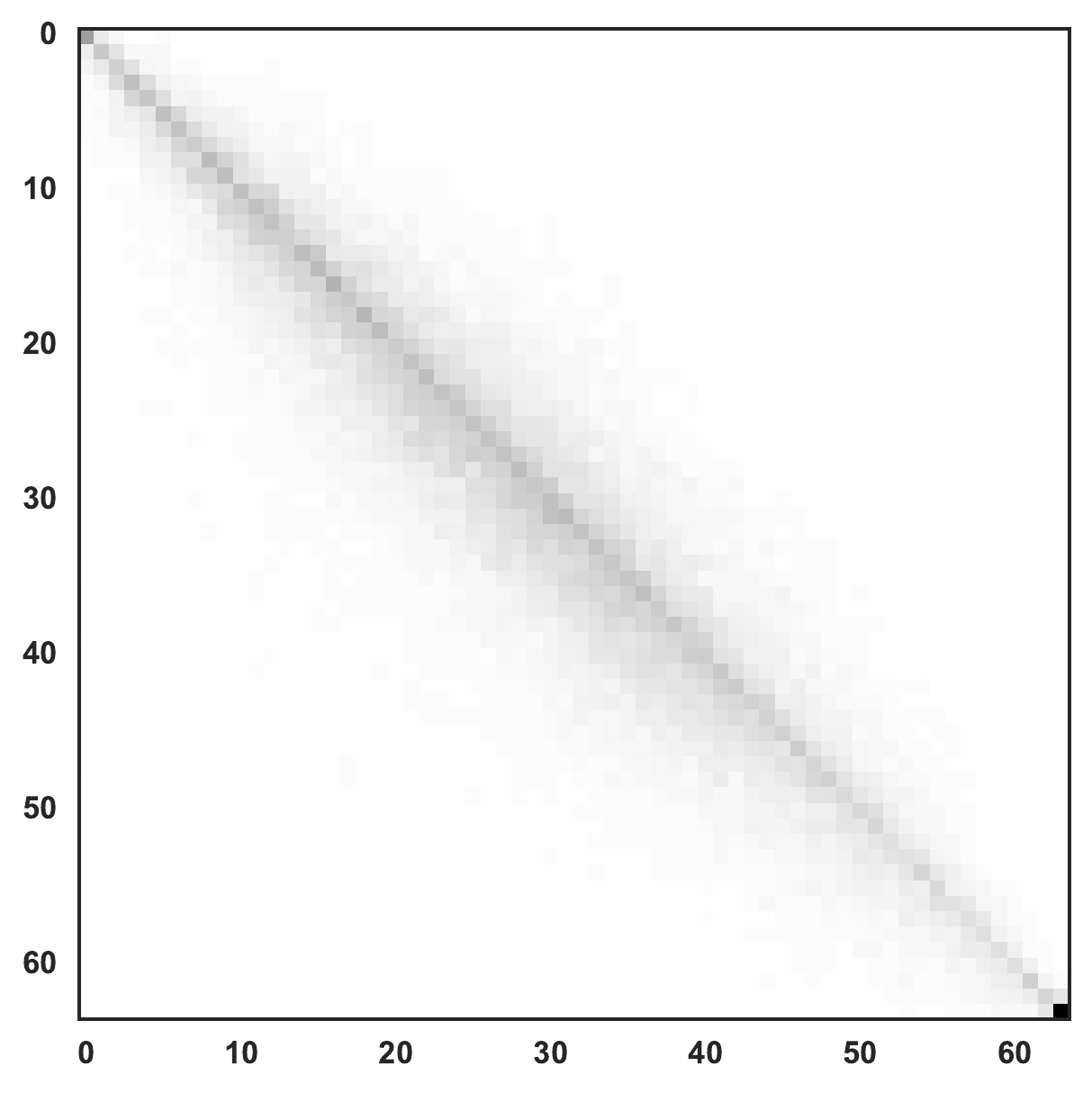}
\caption{CIFAR-10 (Red,Green)}
\end{subfigure}
\begin{subfigure}[t]{0.32\textwidth}
\includegraphics[width=\textwidth]{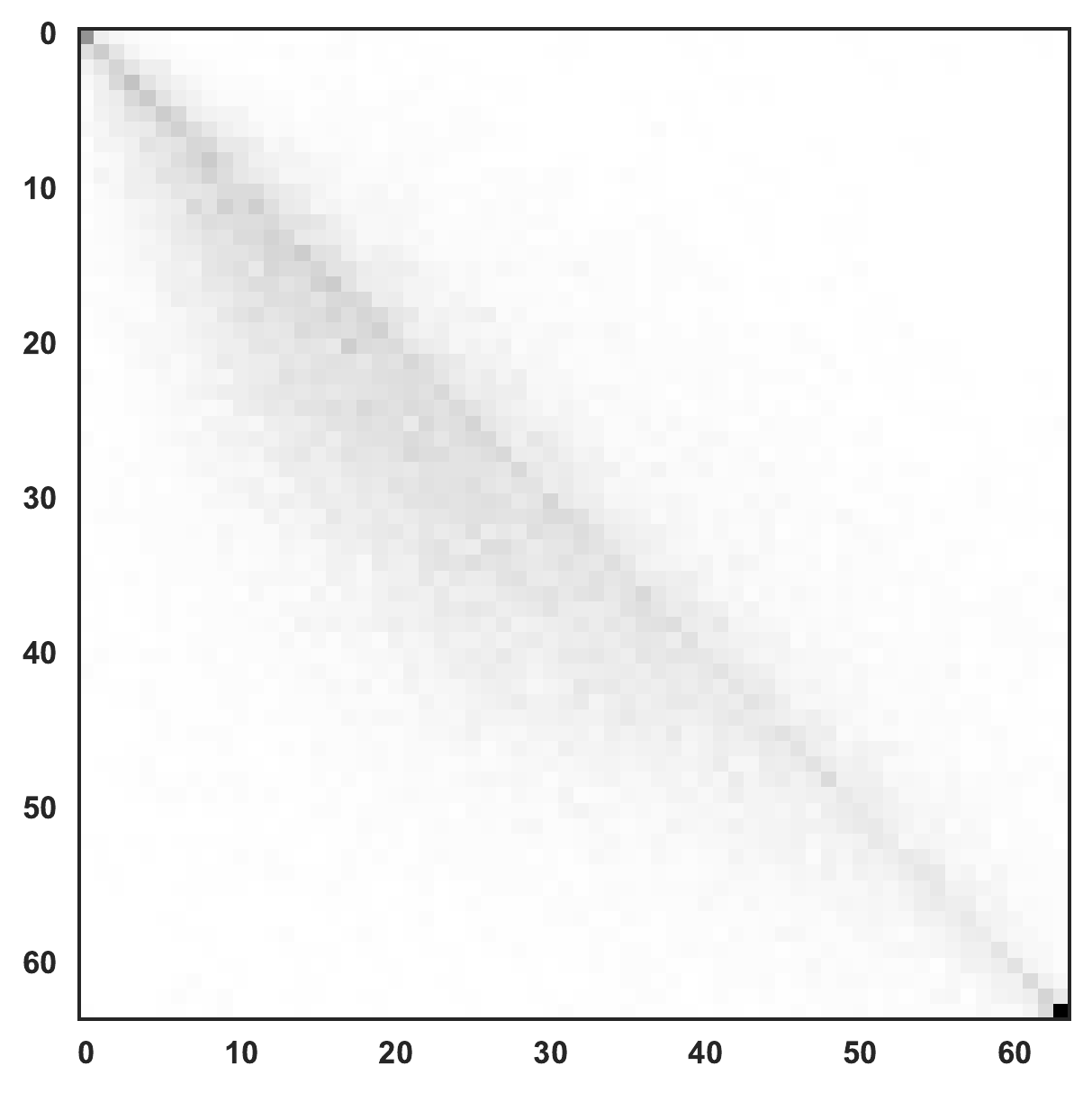}
\caption{CIFAR-10 (Red,Blue)}
\end{subfigure}
\begin{subfigure}[t]{0.32\textwidth}
\includegraphics[width=\textwidth]{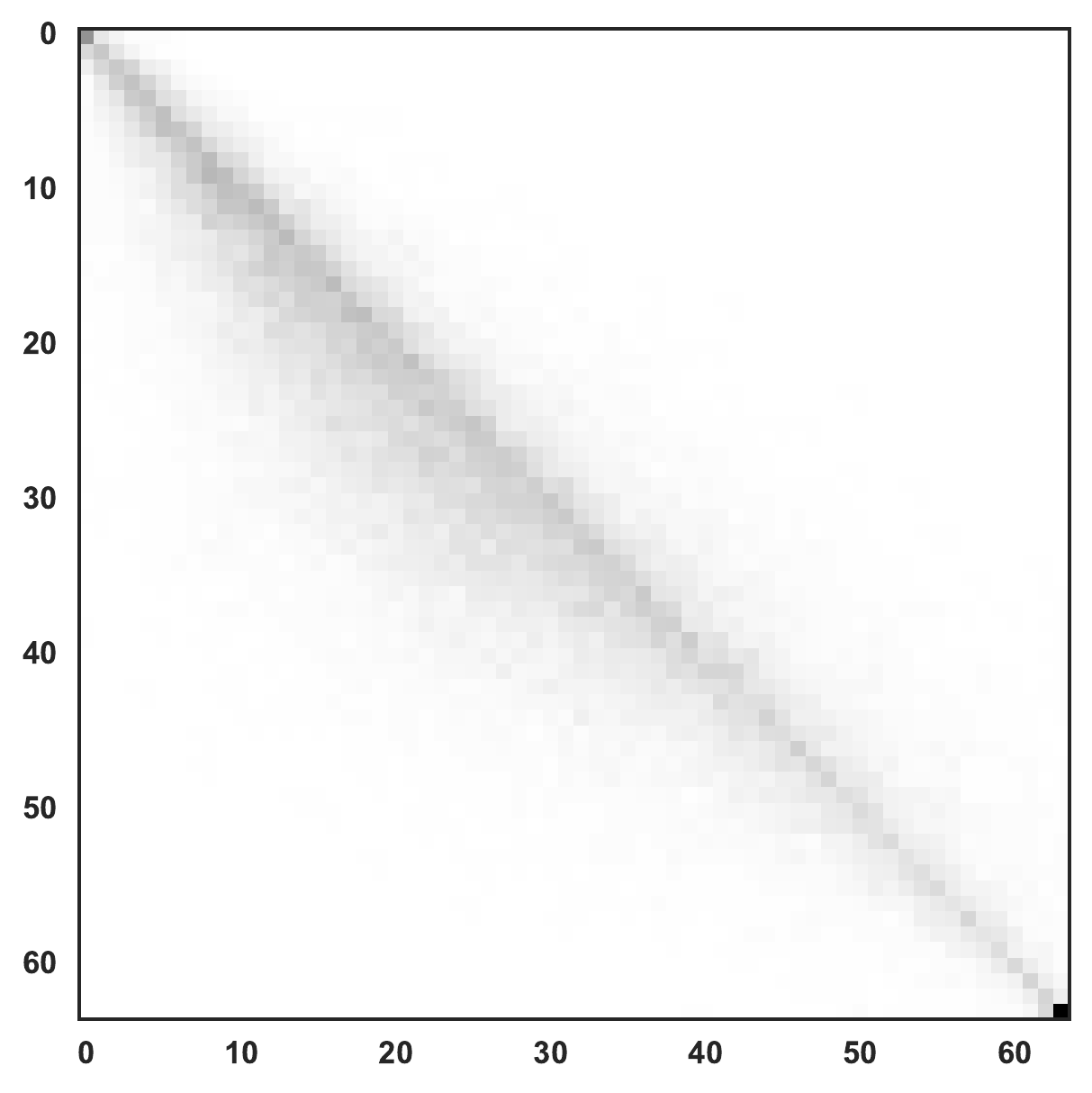}
\caption{CIFAR-10 (Green,Blue)}
\end{subfigure}
\caption{\label{fig:marginals} Marginal distributions of the datasets from Section \ref{subsec:experiments:realworld}. For the 3d datasets, we show the three 2d views of the data.}
\end{figure*}

\end{document}